\PassOptionsToPackage{table}{xcolor}
\documentclass{article} 
\usepackage[T1]{fontenc}
\usepackage{times}
\usepackage{iclr2027_conference}

\usepackage{amsmath}
\usepackage{amssymb}
\usepackage{amsfonts}
\usepackage{bm}

\usepackage{amsmath,amsfonts,bm}

\def\eqref#1{equation~\ref{#1}}

\def\1{\bm{1}}

\DeclareMathAlphabet{\mathsfit}{\encodingdefault}{\sfdefault}{m}{sl}
\SetMathAlphabet{\mathsfit}{bold}{\encodingdefault}{\sfdefault}{bx}{n}

\def\gA{{\mathcal{A}}}

\def\gG{{\mathcal{G}}}

\def\gM{{\mathcal{M}}}

\def\sV{{\mathbb{V}}}

\providecommand{\gM}{{\mathcal{M}}}
\providecommand{\gG}{{\mathcal{G}}}
\providecommand{\gA}{{\mathcal{A}}}

\providecommand{\sV}{{\mathbb{V}}}

\newcommand{\reas}{r}
\newcommand{\ans}{y}
\newcommand{\traj}{\tau}
\newcommand{\correctness}{z}

\newcommand{\xattn}{r_{\mathrm{attn}}}

\newcommand{\DTClin}{\mathrm{DTC}_{\mathrm{lin}}}
\newcommand{\DTCprod}{\mathrm{DTC}_{\mathrm{prod}}}

\newcommand{\CNSL}{C_{\mathrm{NSL}}}
\newcommand{\Cmean}{C_{\mathrm{mean}}}

\newcommand{\PGattn}{\mathrm{UQAC}_{\mathrm{attn}}}

\newcommand{\PGmean}{\mathrm{UQAC}_{\mathrm{mean}}}

\usepackage{url}
\usepackage{xspace}
\usepackage{booktabs}
\usepackage{graphicx}
\usepackage{enumitem}
\usepackage{subcaption}
\usepackage{wrapfig}
\usepackage{float}
\usepackage{placeins}
\usepackage{needspace}
\usepackage{multirow}
\usepackage{makecell}
\usepackage{array}
\usepackage{tabularx}
\usepackage{xcolor}
\definecolor{boxframe}{RGB}{175,210,240}
\definecolor{boxback}{RGB}{242,242,245}
\usepackage{tcolorbox}
\newtcolorbox{prompt}[1][]{
    colback=boxback,
    colframe=boxframe,
    colbacktitle=boxframe,
    coltitle=white,
    fontupper=\footnotesize,
    boxsep=5pt,
    left=2pt,
    right=2pt,
    top=2pt,
    bottom=2pt,
    boxrule=1pt,
    #1,
}
\usepackage{arydshln}
\usepackage[pagebackref=false,colorlinks,citecolor=brown,linkcolor=red,urlcolor=blue]{hyperref}
\usepackage[capitalise]{cleveref}
\Crefname{appendix}{Appendix}{Appendices}
\crefname{appendix}{Appendix}{Appendices}
\Crefname{subappendix}{Appendix}{Appendix}
\crefname{subappendix}{Appendix}{Appendix}
\Crefname{subsubappendix}{Appendix}{Appendix}
\crefname{subsubappendix}{Appendix}{Appendix}
\AddToHook{cmd/appendix/after}{%
  \crefalias{section}{appendix}%
  \crefalias{subsection}{subappendix}%
  \crefalias{subsubsection}{subsubappendix}%
}

\newcommand{\ours}{\textsc{DTC}\xspace}

\title{Probability is Not Enough: Exploring and Counting Divergent Tokens for Reasoning Uncertainty Quantification in LLMs}

\author{Feiyang Li$^{1,2*}$, Shengjing Liu$^{1*}$, Qi Zhan$^{1}$, Sijie Cheng$^{2,3}$,Weiqing Wang$^{1}$, \\
\textbf{Hongwen Chen$^{1}$, Yuxuan Yang$^{1}$, Wen Wang$^{2,4}$, Yile Wang$^{1\dagger}$, Hui Huang$^{1}$} \\
\mdseries $^1$College of Computer Science and Software Engineering, Shenzhen University \ \ $^2$RayNeo.AI \\
$^3$Tsinghua University \ \ $^4$Behavioral and Spatial AI Lab, Peking University \& Tongji University \\
\texttt{lfy20040214@gmail.com} \ \ \ \ \texttt{wangyile@szu.edu.cn} \\
}

\iclrfinalcopy
\makeatletter
\def\@maketitle{\vbox{\hsize\textwidth
{\LARGE\sc \@title\par}
\ificlrfinal
    \lhead{}%
    \def\And{\end{tabular}\hfil\linebreak[0]\hfil
            \begin{tabular}[t]{l}\bf\rule{\z@}{24pt}\ignorespaces}%
    \def\AND{\end{tabular}\hfil\linebreak[4]\hfil
            \begin{tabular}[t]{l}\bf\rule{\z@}{24pt}\ignorespaces}%
    \begin{tabular}[t]{l}\bf\rule{\z@}{24pt}\@author\end{tabular}%
\else
    \lhead{Under review as a conference paper at ICLR 2027}%
    \def\And{\end{tabular}\hfil\linebreak[0]\hfil
            \begin{tabular}[t]{l}\bf\rule{\z@}{24pt}\ignorespaces}%
    \def\AND{\end{tabular}\hfil\linebreak[4]\hfil
            \begin{tabular}[t]{l}\bf\rule{\z@}{24pt}\ignorespaces}%
    \begin{tabular}[t]{l}\bf\rule{\z@}{24pt}Anonymous authors\\Paper under double-blind review\end{tabular}%
\fi
\vskip 0.3in minus 0.1in}}
\makeatother
\begin{document}

\ificlrfinal
\fancypagestyle{arxivfirst}{%
  \fancyhf{}%
  \fancyhead[L]{\includegraphics[height=18pt]{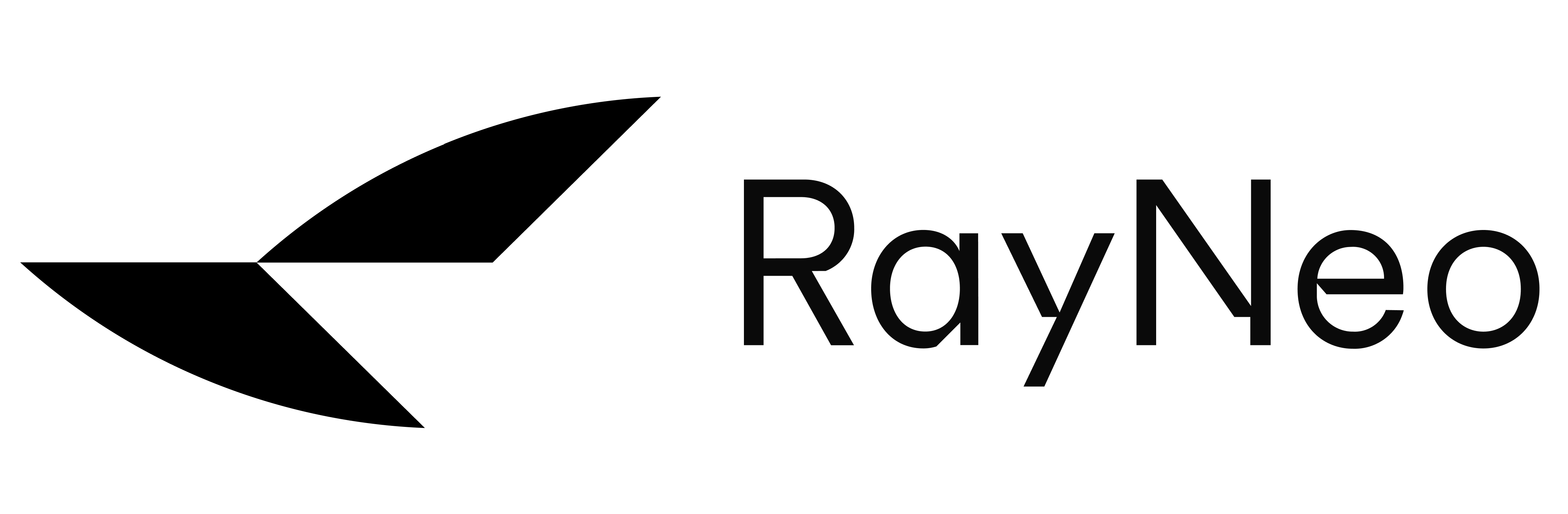}}%
  \fancyfoot[C]{\thepage}%
  \renewcommand{\headrulewidth}{1pt}%
  \renewcommand{\footrulewidth}{0pt}%
  \renewcommand{\headrule}{{\color{gray}\hrule height \headrulewidth width \headwidth \vskip-\headrulewidth}}%
}
\fancyhf{}
\fancyfoot[C]{\thepage}
\renewcommand{\headrulewidth}{1pt}
\renewcommand{\footrulewidth}{0pt}
\renewcommand{\headrule}{{\color{gray}\hrule height \headrulewidth width \headwidth \vskip-\headrulewidth}}
\maketitle
\thispagestyle{arxivfirst}
\renewcommand{\thefootnote}{\fnsymbol{footnote}}
\footnotetext[1]{Equal contribution.}
\footnotetext[2]{Corresponding author.}
\renewcommand{\thefootnote}{\arabic{footnote}}
\setcounter{footnote}{0}
\else
\maketitle
\thispagestyle{fancy}
\setcounter{footnote}{0}
\fi

\begin{abstract}
As the chain-of-thought reasoning capabilities of large language models improve, evaluating and calibrating their reasoning confidence is becoming increasingly important for quantifying the uncertainty of their answers. Current methods for estimating the confidence of large language models are generally based on probabilities of selected key tokens, but the underlying mechanism remains unclear. Our pilot study finds that replacing selected token probabilities with coarse substitutes can also improve calibration, motivating us to further explore effective signals of model confidence. We introduce Divergent Token Confidence (DTC), a framework that estimates confidence by counting tokens at which two models strongly disagree during decoding. DTC identifies these divergent tokens using the Jensen--Shannon divergence between next-token distributions evaluated along the same reasoning trajectory. We find that their count is almost negatively associated with answer accuracy, thereby can serve as a simple yet effective signal for uncertainty quantification. DTC supports both white-box and black-box evaluation using auxiliary models, without explicit training and affecting the generation process. Experiments across multiple model families and six mathematical benchmarks demonstrate improved calibration over the considered probability-based and verbalized-based baselines. Under white-box evaluation, the count-only estimator achieves an average expected calibration error of 13.0\%, compared with 32.7\%–42.4\% for standard full-sequence confidence methods. In black-box settings, it also improves calibration over the original verbalized scores. For example, mean expected calibration error falls from 32.1\%–40.2\% to 13.7\%–16.3\% on DeepSeek-V3.2. These findings provide new insights for improving reasoning uncertainty quantification in large language models. 
The code is released at https://github.com/szu-tera/DTC.git.
\end{abstract}

\section{Introduction}
\label{sec:intro}

Large language models have shown strong capability to solve complex tasks through Chain-of-Thought (CoT) reasoning~\citep{Wei-2022-CoT, OpenAI-2024-o1, DeepSeek-2025-R1}. 
However, a reasoning trajectory that yields a correct answer may still contain unreliable intermediate decisions, while one that yields an incorrect answer may still be logically reliable~\citep{Wei-2022-CoT, Bao-2025-CoT-Mimic, Landscape-2025-LoT}. 
Therefore, estimating the confidence of a model’s reasoning trajectory is becoming increasingly important. First, a mismatch between model confidence and its answers can amplify the model’s unreliability, thereby limiting its deployment in safety-critical scenarios~\citep{Clusmann-2023-LLM-Medicine}. 
Second, understanding reasoning confidence can help us determine whether we need to rely on the outputs of models. When a reasoning path is identified as unreliable, the model can abstain from answering~\citep{Madhusudhan-2025-Abstention}, or it can be deferred to human review~\citep{Devic-2025-Human-Centered-UQ}. 
Finally, it has been shown that confidence itself can also be leveraged to improve the model’s own reasoning performance~\citep{fu2026deep}.

We refer to this confidence estimation task as \emph{Reasoning Uncertainty Quantification}: assigning a confidence score to each query and its reasoning path to reflect the reliability of the path and its final answer~\citep{Liu-2025-UQ-Survey, Zhang-2025-CoTUQ}.
Existing methods typically aggregate token probabilities over the entire trajectory, but the resulting scores can be overconfident~\citep{Orgad-2025-Know-More, Li-2025-UQAC, Zhang-2025-CoTUQ}.
To mitigate this overconfidence, \citet{Li-2025-UQAC} propose Uncertainty Quantification with Attention Chain (UQAC), which selects answer-related CoT tokens and multiplies their probabilities to obtain a path-level confidence score.
This design suggests that the probabilities of answer-related tokens provide useful information for estimating confidence in answer correctness.
However, its product score depends jointly on these probabilities and the selected-token count, leaving it unclear whether the calibration gains come from the selected probabilities or the count.

To test whether the selected probabilities are necessary for these calibration gains, we keep UQAC's selected token set fixed and replace the selected probabilities with either the trajectory's mean token probability or a constant shared across trajectories from the same model on a given dataset.
Both replacements generally improve calibration in our case study, suggesting that the selected tokens' specific probabilities may not be necessary for these gains.
In particular, replacing the probabilities with a shared constant makes the score depend only on the number of selected tokens, motivating us to examine token count as a calibration signal~(\Cref{sec:uqac-audit}).

\begin{figure}[t]
  \centering
  \vspace*{11pt}%
  \makebox[\linewidth][c]{%
    \includegraphics[width=\linewidth,keepaspectratio,trim=0bp 223bp 137bp 1bp,clip]{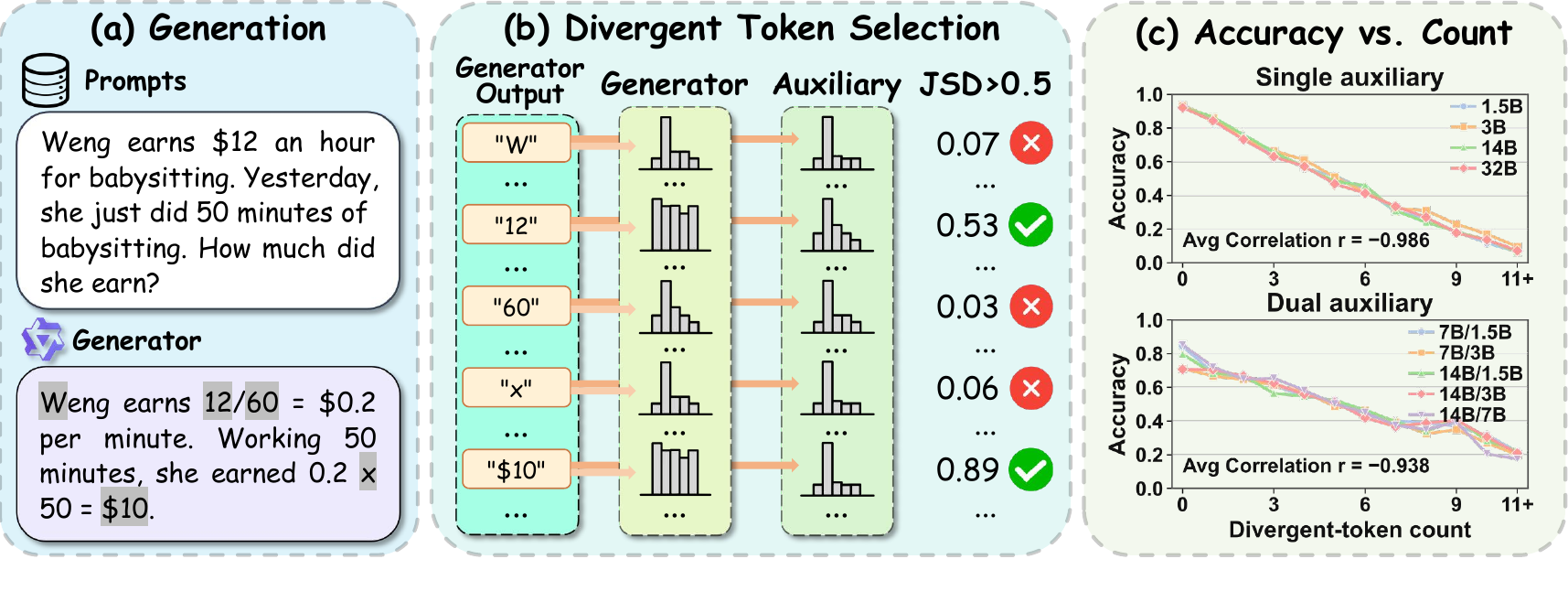}%
  }
  \caption{Divergent-token selection and its relationship to path accuracy. (a) A generator produces a reasoning trajectory. (b) A token is selected when the JSD between two models' next-token distributions exceeds \(\theta\). (c) Path accuracy decreases as the number of divergent tokens increases under single- and dual-auxiliary probing.}
  \label{fig:divergent-token-count-vs-acc}
\end{figure}

To obtain the token count signal that reflects reasoning path reliability, we draw on inter-model disagreement as an uncertainty signal~\citep{Kruse-2025-MUSE, Sun-2024-CrossCheckGPT}.
We hypothesize that \emph{unreliable reasoning paths contain more tokens at which models strongly disagree}.
We measure token-level disagreement using the Jensen--Shannon divergence (JSD) between two models' next-token distributions and define tokens whose divergence exceeds a threshold \(\theta\) as divergent tokens.
Their count summarizes the frequency of strong disagreement along the reasoning path and is negatively associated with path accuracy in our experiments~(\Cref{fig:divergent-token-count-vs-acc}).

We therefore introduce \textbf{Divergent Token Confidence (DTC)}, which converts divergent-token count into path-level confidence. 
DTC provides two estimators: \(\DTClin\) maps the count directly to confidence, while \(\DTCprod\) combines the count with a full-sequence probability score. 
The same procedure can be applied in white-box settings using the generator's information and in black-box settings using auxiliary models on the generated trajectory, without requiring access to the generator's logits. 
Across multiple model families and mathematical benchmarks, DTC improves calibration over full-sequence likelihood-based, verbalized-confidence, and UQAC baselines~(\Cref{sec:rq2,sec:rq3}). 
DTC also reduces overconfidence in verbalized scores on existing trajectories~(\Cref{sec:verbalized-dtc}).

\section{Related Work}
\label{sec:related_work}

\paragraph{Probability-based confidence and verbalized confidence.}
In white-box uncertainty quantification, token probabilities or predictive uncertainty from the generating model are commonly used to construct response-level confidence scores.
Sequence-level uncertainty estimators typically aggregate token-level signals over a generated response~\citep{Malinin-2021-Predictive-Entropy}.
Relevance-aware methods account for differences in how tokens contribute to meaning or the final answer, weighting these signals by token relevance, sentence relevance, or contextual information~\citep{Duan-2024-Predictive-Uncertainty, Bakman-2024-MARS, Lin-2024-Contextualized-Sequence-Likelihood}.
For responses that include CoT reasoning, CoT-UQ and UQAC further use the relevance of tokens in the reasoning chain to the final answer to estimate response-level uncertainty~\citep{Zhang-2025-CoTUQ, Li-2025-UQAC}.
In black-box settings, verbalized confidence is a common approach to confidence estimation~\citep{Wang-2026-BlackBox-UE}.
Such methods either elicit confidence through additional prompts after a response has been generated or ask the model to report confidence or a distribution over candidate answers alongside its CoT and final answer~\citep{Tian-2023-Just-Ask, Wang-2025-Verbalized-Distribution}.
The latter couples reasoning with confidence elicitation: the choice of prompt may alter the CoT trajectory and lead to lower accuracy or overconfidence in mathematical reasoning~\citep{Wang-2025-Verbalized-Distribution}.
Our work reveals their limitations and investigates the novel divergent-token count as a confidence signal in both white-box and black-box settings.

\paragraph{Cross-model signals for confidence estimation.}
Another line of work uses multiple models, base models, or model perturbations to estimate the reliability of generated responses.
MUSE and CrossCheckGPT use multi-model consensus or cross-system consistency to improve confidence estimation~\citep{Kruse-2025-MUSE, Sun-2024-CrossCheckGPT}.
BaseCal uses signals from base models to calibrate the confidence of post-trained models~\citep{Tan-2026-BaseCal}.
More closely related to our work are methods that estimate reasoning uncertainty from token entropy.
Some aggregate the generating model's token entropy along a CoT trajectory; others use randomly perturbed models or additional models and aggregate their token entropy along the same trajectory to estimate uncertainty over the reasoning trajectory~\citep{Zhang-2026-TokUR, Gorbett-2026-Cross-Model}.
These studies primarily focus on response-level uncertainty or its ranking performance, whereas our focus is on a bounded score that can be directly interpreted as confidence and used for calibration.


\section{Preliminaries and Pilot Study}
\label{sec:background}

\subsection{Reasoning Uncertainty and Calibration}
\label{sec:reas-calib}

The \textit{prediction uncertainty} (or query uncertainty) \(U(x)\) characterizes uncertainty over answers to a query \(x\) before conditioning on a particular realized reasoning path, and is often estimated by sampling multiple answers to the same query~\citep{Kuhn-2023-Semantic-Uncertainty}. In this work, we study \emph{reasoning uncertainty} \(U(x,\reas)\), the uncertainty in final-answer reliability conditioned on both the query \(x\) and the realized reasoning process \(\reas\)~\citep{Liu-2025-UQ-Survey, Zhang-2025-CoTUQ}.

Given the query \(x\), a generating model \(\gG\) produces a trajectory \(\traj=(\traj_1,\ldots,\traj_T)\) of \(T\) tokens, consisting of a CoT path \(\reas\) and a final answer \(\ans\); a reference answer or evaluator provides the binary correctness label \(\correctness\in\{0,1\}\).
We operationalize \(U(x,\reas)\) through a calibrated confidence \(C(\ans, x,\reas)\in[0,1]\) that estimates \(P(\correctness{=}1\mid x,\reas,\ans)\), with higher confidence indicating lower reasoning uncertainty.
Uncertainty is perfectly calibrated if
\begin{equation}
  P\bigl(\correctness{=}1 \mid \ans, x,\reas \bigr) = q, \text{ s.t. } C(\ans, x,\reas) = q.
  \label{eq:calibration}
\end{equation}

Expected calibration error (ECE;~\citealp{Guo-2017-Calibration}) is used to quantify the degree of miscalibration: we partitions \(N\) scored trajectories into \(M\) equal-width confidence bins \(\{B_1,\ldots,B_M\}\) and then calculates the absolute gap between average accuracy \(\mathrm{acc}(B_m)\) and confidence \(\mathrm{conf}(B_m)\):
\begin{equation}
  \mathrm{ECE}
  \;=\;
  \sum_{m=1}^{M}\frac{|B_m|}{N}\,
  \bigl|\mathrm{acc}(B_m)-\mathrm{conf}(B_m)\bigr|.
  \label{eq:ece}
\end{equation}
Lower ECE indicates better calibration. Unless noted, we report ECE with \(M{=}20\).

\subsection{Sequence Confidence from Token Probabilities}
\label{sec:semantic-token-uq}

Standard full-sequence confidence scores aggregate token probabilities over the entire trajectory. One such score is the length-normalized sequence likelihood (NSL;~\citealp{Malinin-2021-Predictive-Entropy}).
Given the generator's token probabilities \(p_t=P_{\gG}(\traj_t\mid x,\traj_{<t})\), we have
\begin{equation}
  C_{\mathrm{NSL}}(\traj, x)
  \;=\;
  \Bigl(\prod_{t=1}^{T} p_t\Bigr)^{1/T}
  \;=\;
  \exp\!\Bigl(\frac{1}{T}\sum_{t=1}^{T}\log p_t\Bigr).
  \label{eq:nsl}
\end{equation}
Another similar full-sequence confidence score is the mean token probability~\citep{Orgad-2025-Know-More}, denoted as \(\Cmean=\frac{1}{T}\sum_{t=1}^{T}p_t\), which could be overconfident~\citep{Zhang-2025-CoTUQ}.

Recent work argue that not every token is equally diagnostic of answer correctness and accordingly assigns each token a relevance weight \(w_t\) to form a relevance-weighted score~\citep{Bakman-2024-MARS}
\begin{equation}
  C_{\mathrm{rel}}(\traj, x)
  \;=\;
  \prod_{t=1}^{T} p_t^{\,w_t}
  \;=\;
  \exp\!\Bigl(\sum_{t=1}^{T} w_t\log p_t\Bigr).
  \label{eq:rel-weighted}
\end{equation}
Closest to our setting, UQAC instantiates this idea on CoT \(\reas\) through hard selection. It constructs an attention chain \(\xattn\) by backtracking from \(\ans\) with attention weights, optionally refining the chain by similarity to the answer.
Let \(\mathcal{S}\) denote the selected-token set induced by this chain (typically including answer tokens) and \(S=|\mathcal{S}|\) its size.
In the generic formulation above, this hard selection corresponds to \(w_t=\mathbf{1}\{t\in\mathcal{S}\}\).
The primary score is the unnormalized product
\begin{equation}
  \PGattn=\prod_{t\in\mathcal{S}} p_t,.
  \label{eq:uqac-prod}
\end{equation}
Although UQAC is effective to a certain extent, the product depends jointly on these probabilities and the number of selected-token $|\mathcal{S}|$. This makes it unclear whether the probability is truly effective, and the impact of the number of selected tokens is also unclear. Therefore, the calibration source needs further investigation and we are motivated to test whether the specific selected-token probabilities are truly necessary for calibration gains and to explore better calibration signals. 

\subsection{Do Selected-Token Probabilities Explain UQAC's Calibration Gains?}
\label{sec:uqac-audit}

\setlength{\abovecaptionskip}{2pt}%
\setlength{\belowcaptionskip}{0pt}%
\noindent
\begin{minipage}[t]{0.66\textwidth}
  \vspace{0pt}%
  \setlength{\parindent}{1.2em}%
  \noindent
  We keep UQAC's selected-token set \(\mathcal{S}\) fixed for each trajectory and construct two variants that progressively remove specific probability information.
  \(C_{\mathrm{mean}}^{S}\) replaces each selected-token probability with the trajectory's mean token probability \(\Cmean\) before taking their product, as defined in~\Cref{sec:semantic-token-uq}. \(C_{\mathrm{global}}^{S}\) further replaces the trajectory-specific mean with \(C_{\mathrm{global}}\), obtained by averaging \(C_{\mathrm{mean}}\) over all trajectories with the same model on a given dataset. This value is shared across those trajectories, so only the selected-token count \(S\) varies.
  \Cref{fig:uqac-ece} compares these two variants, the original \(C_{\mathrm{mean}}\), and UQAC across Qwen2.5 family.
  Although \(C_{\mathrm{mean}}\) itself has high ECE, the resulting \(C_{\mathrm{mean}}^{S}\) achieves substantially lower ECE than UQAC.
  Even with a fixed constant for all trajectories from each model, \(C_{\mathrm{global}}^{S}\) generally remains better calibrated than UQAC.  Given the well calibration performance of \(C_{\mathrm{mean}}^{S}\), we first propose UQAC\(_{\mathrm{mean}}\) as an improved variant of UQAC, and the results in main experiments (Table~\ref{tab:whitebox-ece}) indeed demonstrate its advantages. 
\end{minipage}\hfill
\begin{minipage}[t]{0.32\textwidth}
  \vspace{0pt}%
  \centering
  \includegraphics[width=\linewidth]{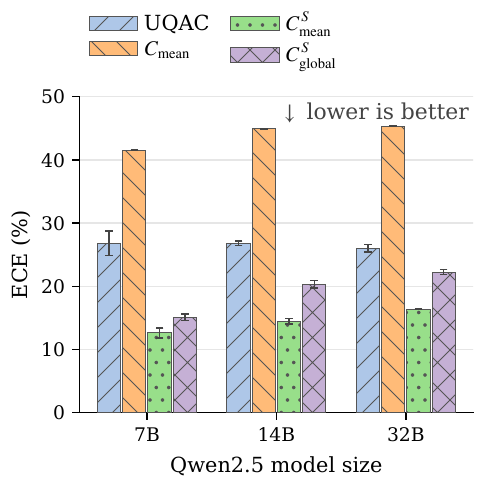}
  \captionof{figure}{Using the selected count as an exponent yields the lowest ECE across model sizes.}
  \label{fig:uqac-ece}
\end{minipage}

\vspace{0.15em}
More importantly, this pilot study shows that replacing the selected tokens' probabilities generally improves calibration over UQAC, thus the gains may not depend on the specific value of probabilities. When probabilities are replaced with a shared constant, the score depends only on the number of selected tokens, suggesting that token count may serve as a alternative calibration signal.

\vspace{0.25em}

\subsection{Motivation: Counting Divergent Tokens}
\label{sec:motivation}
\label{sec:method-selection}

To obtain such a count for calibration, we draw on inter-model disagreement as an uncertainty signal~\citep{Lakshminarayanan-2017-Deep-Ensembles, Sun-2024-CrossCheckGPT, Kruse-2025-MUSE}.
We hypothesize that \emph{unreliable reasoning paths contain more tokens at which models strongly disagree}.
Specifically, we use teacher forcing to measure disagreement between two models' next-token distributions at each position along the same fixed trajectory.

At reasoning position \(t\), let \(P_{\gM}(\cdot\mid c_t)\) denote model \(\gM\)'s next-token distribution under the shared prefix \(c_t=(x,\traj_{<t})\). Denoting two distinct distributions as \(P_t\) and \(Q_t\), 
we use Jensen--Shannon divergence (JSD) to measure their disagreement for the token \(\traj_t\):
\begin{equation}
  \mathrm{JSD}(P_t,Q_t)
  \;=\;
  \mathbb{H}\!\left(\frac{P_t+Q_t}{2}\right)
  -\tfrac12\,\mathbb{H}(P_t)
  -\tfrac12\,\mathbb{H}(Q_t),
  \label{eq:js}
\end{equation}
where \(\mathbb{H}(\cdot)\) is Shannon entropy.
Models in different families disagree to different degrees, so we set the threshold \(\theta\) separately for each family on a small validation set.
We define a token as a \textit{divergent token} if \(\mathrm{JSD}(P_t,Q_t) > \theta\) and set of divergent token positions along a reasoning path is:
\begin{equation}
  T_{\rm div}(\theta)
  \;=\;
  \bigl\{\,t\in\{1,\dots,T\}:\mathrm{JSD}(P_t,Q_t) > \theta\,\bigr\}.
  \label{eq:divergent-token-set}
\end{equation}
We use JSD for its symmetry and boundedness and see \Cref{app:divergence-ablation} for discussion on alternative measures of token-level disagreement.

Based on the definition, we can first measure the confidence of generator \(\gG\) with an auxiliary model \(\gA\) in the \emph{single-auxiliary} setting, which applies when \(\gG\) is a white-box model. When the target model is a black-box model and we cannot access its token distribution, we assume that \emph{the relationship between the number of divergent tokens and reasoning-path reliability remains reasonably stable across different model pairs}, which allow us to calibrate in a \emph{dual-auxiliary} setting by comparing two external auxiliary models from the same family, denoted as \(\gA'\) and \(\gA''\).



To verify that the number of divergent tokens \(\lvert T_{\rm div}(\theta)\rvert\) can serve as a signal of model uncertainty, we first analyze the relationship between answer accuracy against the number of divergent tokens for different model pairs. The results are shown in \Cref{fig:divergent-token-count-vs-acc}(c). In both single- and dual-auxiliary settings, accuracy decreases nearly monotonically as \(\lvert T_{\rm div}(\theta)\rvert\) increases, supporting that the number of divergent tokens can reveal reasoning-path reliability to some extent.
\section{\ours: Divergent Token Confidence}
\label{sec:method}
\label{sec:method-estimators}

Given the divergent-token count \(m=\lvert T_{\rm div}(\theta)\rvert\) defined in~\Cref{eq:divergent-token-set}, \ours constructs two path-level confidence estimates.
The primary estimator \(\DTClin\) maps \(m\) directly to confidence, while \(\DTCprod\) uses \(m\) to recalibrate the standard full-sequence confidence \(\Cmean\) from~\Cref{sec:semantic-token-uq}.

\paragraph{\boldmath\(\DTClin\): Count-linear confidence.}
Motivated by the decrease in accuracy with increasing divergent token count~(\Cref{fig:divergent-token-count-vs-acc}(b,c)), we define
\begin{equation}
  \DTClin(m)=
  \begin{cases}
    a-\dfrac{a-b}{n}\,m, & 0\le m<n,\\[4pt]
    b, & m\ge n.
  \end{cases}
  \label{eq:clin}
\end{equation}
Confidence starts at \(a\), decreases linearly with \(m\), and reaches a floor of \(b\) at \(m=n\).
We use \(a=0.95\), \(b=0.05\), and \(n=10\) in all main experiments; sensitivity to \(n\) is examined in~\Cref{app:clin-n-ablation}.
Once the divergent tokens are selected, this mapping depends only on their count.

\paragraph{\boldmath\(\DTCprod\): Trajectory-mean product confidence.}
\Cref{sec:uqac-audit} shows that replacing selected-token probabilities with the trajectory mean \(\Cmean\) and taking their product reduces overconfidence. We therefore combine \(\Cmean\) with the uncertainty-informed count \(m\) for confidence estimation:
\begin{equation}
  \DTCprod(m)=\Cmean^{\,m+k}.
  \label{eq:pprod}
\end{equation}
For \(\Cmean\), we use probabilities from the generator in white-box settings~(\Cref{sec:whitebox-setup}) and the larger auxiliary on the same frozen trajectory in black-box settings~(\Cref{sec:blackbox-setup}).
We use \(k=4\) unless otherwise specified, avoiding a score of one solely due to a zero count and sensitivity to \(k\) is examined in~\Cref{app:dtcprod-k-ablation}.
When \(\Cmean\in(0,1)\), a larger \(m\) yields a lower score. Compared with \(\DTClin\), \(\DTCprod\) further uses the trajectory-level probability \(\Cmean\), giving finer-grained confidence to trajectories that share the same \(m\).
The uncertainty-informed count \(m\) can also be combined with other overconfident scores to improve calibration~(\Cref{sec:analysis}).

\par
\begin{table}[!t]
  \centering
  \footnotesize
  \setlength{\tabcolsep}{2.0pt}
  \renewcommand{\arraystretch}{1.05}
  \caption{Uncertainty quantification performance (ECE) in white-box settings.
  In each row, the best result is in \textbf{bold} and the second-best one
  is \underline{underlined} (excluding PRM).
  \dag : UQAC variant by ours.}
  \label{tab:whitebox-ece}
  \ifcsname wbeceleftbox\endcsname\else\newsavebox{\wbeceleftbox}\fi
  \ifcsname wbecerightbox\endcsname\else\newsavebox{\wbecerightbox}\fi
  \sbox{\wbeceleftbox}{%
  \begin{tabular}[t]{@{}wl{5.6em}wc{2.3em}wc{3.55em}wc{3.55em}wc{3.55em}wc{3.55em}wc{3.55em}wc{3.55em}wc{3.55em}wc{3.55em}wc{3.55em}@{}}
    \toprule
    \multirow{2}{*}{\makecell[l]{\textbf{Reasoning} \\ \textbf{Models}}} & \multirow{2}{*}{\centering\textbf{Acc}} & \multirow{2}{*}{\centering{\boldmath$\CNSL$}} & \multirow{2}{*}{\centering{\boldmath$\Cmean$}} & \multirow{2}{*}{\centering\makecell[c]{\textbf{Entropy}\\\textbf{Conf.}}} & \multirow{2}{*}{\centering\textbf{BaseCal}} & \multicolumn{2}{c}{\textbf{UQAC}} & \multirow{2}{*}{\centering\textbf{Verb.}} & \multicolumn{2}{c}{\textbf{DTC (Ours)}} \\
    \cmidrule(lr){7-8}\cmidrule(lr){10-11}
    & & & & & & \textbf{attn} & \textbf{mean}\textsuperscript{\dag} & & \textbf{prod} & \textbf{lin} \\
    \midrule
    \multicolumn{11}{c}{\textit{MATH-500}} \\
    \midrule
    Qwen2.5\mbox{-}7B & 76.1 & \cellcolor[RGB]{231,239,247}43.6 & \cellcolor[RGB]{235,241,248}45.3 & \cellcolor[RGB]{207,224,241}32.8 & \cellcolor[RGB]{231,239,247}43.7 & \cellcolor[RGB]{193,215,237}26.8 & \cellcolor[RGB]{162,195,229}\textbf{12.7} & \cellcolor[RGB]{232,240,247}44.2 & \cellcolor[RGB]{178,206,233}20.0 & \cellcolor[RGB]{163,196,230}\underline{13.4} \\
    Qwen2.5\mbox{-}14B & 80.0 & \cellcolor[RGB]{230,238,246}43.0 & \cellcolor[RGB]{234,241,247}44.9 & \cellcolor[RGB]{220,232,244}38.9 & \cellcolor[RGB]{230,238,246}43.0 & \cellcolor[RGB]{193,215,237}26.8 & \cellcolor[RGB]{165,197,230}\underline{14.2} & \cellcolor[RGB]{211,226,242}34.7 & \cellcolor[RGB]{169,200,231}16.2 & \cellcolor[RGB]{155,192,228}\phantom{0}\textbf{9.9} \\
    Qwen2.5\mbox{-}32B & 82.4 & \cellcolor[RGB]{231,239,247}43.7 & \cellcolor[RGB]{235,241,248}45.4 & \cellcolor[RGB]{223,234,245}39.9 & \cellcolor[RGB]{231,239,247}43.8 & \cellcolor[RGB]{192,214,237}26.1 & \cellcolor[RGB]{170,200,231}\underline{16.2} & \cellcolor[RGB]{226,236,246}41.6 & \cellcolor[RGB]{171,201,232}16.9 & \cellcolor[RGB]{148,187,226}\phantom{0}\textbf{6.6} \\
    Qwen3\mbox{-}8B & 83.9 & \cellcolor[RGB]{219,231,244}38.1 & \cellcolor[RGB]{226,236,245}41.5 & \cellcolor[RGB]{204,222,240}31.8 & \cellcolor[RGB]{215,229,243}36.7 & \cellcolor[RGB]{216,230,243}37.1 & \cellcolor[RGB]{218,231,243}37.8 & \cellcolor[RGB]{240,245,249}47.8 & \cellcolor[RGB]{151,189,226}\phantom{0}\textbf{7.9} & \cellcolor[RGB]{168,200,231}\underline{15.6} \\
    Qwen3\mbox{-}14B & 86.5 & \cellcolor[RGB]{216,229,243}36.8 & \cellcolor[RGB]{224,235,245}40.6 & \cellcolor[RGB]{200,220,239}29.9 & \cellcolor[RGB]{214,228,242}36.2 & \cellcolor[RGB]{194,216,237}27.1 & \cellcolor[RGB]{163,196,229}13.2 & \cellcolor[RGB]{228,237,246}42.5 & \cellcolor[RGB]{146,185,225}\phantom{0}\textbf{5.5} & \cellcolor[RGB]{159,194,229}\underline{11.6} \\
    Qwen3\mbox{-}32B & 83.7 & \cellcolor[RGB]{214,228,242}36.1 & \cellcolor[RGB]{223,234,245}40.2 & \cellcolor[RGB]{199,219,239}29.4 & --- & \cellcolor[RGB]{220,232,244}38.8 & \cellcolor[RGB]{220,232,244}38.6 & \cellcolor[RGB]{216,230,243}37.0 & \cellcolor[RGB]{147,186,225}\phantom{0}\textbf{6.1} & \cellcolor[RGB]{158,193,228}\underline{11.0} \\
    Gemma3\mbox{-}12B & 84.9 & \cellcolor[RGB]{226,236,245}41.5 & \cellcolor[RGB]{232,239,247}44.0 & \cellcolor[RGB]{217,230,243}37.3 & \cellcolor[RGB]{211,226,242}34.8 & \cellcolor[RGB]{209,225,241}33.9 & \cellcolor[RGB]{199,219,239}29.3 & \cellcolor[RGB]{188,212,236}24.3 & \cellcolor[RGB]{167,199,231}\underline{15.2} & \cellcolor[RGB]{164,197,230}\textbf{13.9} \\
    Gemma3\mbox{-}27B & 89.2 & \cellcolor[RGB]{228,237,246}42.3 & \cellcolor[RGB]{233,240,247}44.5 & \cellcolor[RGB]{219,232,244}38.4 & \cellcolor[RGB]{214,228,242}36.0 & \cellcolor[RGB]{169,200,231}16.0 & \cellcolor[RGB]{149,187,226}\phantom{0}\textbf{7.0} & \cellcolor[RGB]{208,224,241}33.4 & \cellcolor[RGB]{167,199,230}15.0 & \cellcolor[RGB]{149,188,226}\phantom{0}\underline{7.1} \\
    \midrule
    \multicolumn{11}{c}{\textit{AMC23}} \\
    \midrule
    Qwen2.5\mbox{-}7B & 53.6 & \cellcolor[RGB]{231,239,247}43.6 & \cellcolor[RGB]{235,241,248}45.3 & \cellcolor[RGB]{204,222,240}31.7 & \cellcolor[RGB]{232,239,247}44.0 & \cellcolor[RGB]{189,212,236}24.7 & \cellcolor[RGB]{152,189,227}\phantom{0}\underline{8.2} & \cellcolor[RGB]{228,237,246}42.5 & \cellcolor[RGB]{168,200,231}15.7 & \cellcolor[RGB]{148,187,226}\phantom{0}\textbf{6.7} \\
    Qwen2.5\mbox{-}14B & 61.2 & \cellcolor[RGB]{229,238,246}42.7 & \cellcolor[RGB]{233,240,247}44.7 & \cellcolor[RGB]{219,232,244}38.5 & \cellcolor[RGB]{230,238,246}43.1 & \cellcolor[RGB]{190,213,236}25.3 & \cellcolor[RGB]{177,205,233}19.7 & \cellcolor[RGB]{212,227,242}35.1 & \cellcolor[RGB]{155,191,228}\phantom{0}\underline{9.7} & \cellcolor[RGB]{155,191,227}\phantom{0}\textbf{9.5} \\
    Qwen2.5\mbox{-}32B & 66.4 & \cellcolor[RGB]{230,238,247}43.3 & \cellcolor[RGB]{234,241,248}45.1 & \cellcolor[RGB]{222,233,244}39.5 & \cellcolor[RGB]{232,239,247}44.0 & \cellcolor[RGB]{189,213,236}24.9 & \cellcolor[RGB]{184,210,235}22.7 & \cellcolor[RGB]{219,231,244}38.3 & \cellcolor[RGB]{155,191,228}\phantom{0}\textbf{9.8} & \cellcolor[RGB]{157,193,228}\underline{10.8} \\
    Qwen3\mbox{-}8B & 68.6 & \cellcolor[RGB]{215,229,243}36.5 & \cellcolor[RGB]{224,234,245}40.5 & \cellcolor[RGB]{199,219,239}29.4 & \cellcolor[RGB]{216,229,243}36.8 & \cellcolor[RGB]{244,247,250}49.4 & \cellcolor[RGB]{220,232,244}38.5 & \cellcolor[RGB]{242,246,250}48.7 & \cellcolor[RGB]{148,187,226}\phantom{0}\textbf{6.8} & \cellcolor[RGB]{154,191,227}\phantom{0}\underline{9.3} \\
    Qwen3\mbox{-}14B & 74.1 & \cellcolor[RGB]{211,226,242}34.6 & \cellcolor[RGB]{221,233,244}39.2 & \cellcolor[RGB]{193,215,237}26.9 & \cellcolor[RGB]{214,228,242}35.9 & \cellcolor[RGB]{194,216,237}27.2 & \cellcolor[RGB]{157,192,228}\textbf{10.5} & \cellcolor[RGB]{212,227,242}35.3 & \cellcolor[RGB]{161,195,229}12.3 & \cellcolor[RGB]{160,194,229}\underline{11.8} \\
    Qwen3\mbox{-}32B & 67.5 & \cellcolor[RGB]{209,225,241}33.7 & \cellcolor[RGB]{219,232,244}38.5 & \cellcolor[RGB]{190,213,236}25.4 & --- & \cellcolor[RGB]{232,239,247}44.0 & \cellcolor[RGB]{218,230,243}37.7 & \cellcolor[RGB]{206,223,240}32.3 & \cellcolor[RGB]{167,199,231}\underline{15.2} & \cellcolor[RGB]{150,188,226}\phantom{0}\textbf{7.5} \\
    Gemma3\mbox{-}12B & 66.8 & \cellcolor[RGB]{225,235,245}40.8 & \cellcolor[RGB]{231,239,247}43.6 & \cellcolor[RGB]{215,229,243}36.4 & \cellcolor[RGB]{212,227,242}35.3 & \cellcolor[RGB]{211,226,242}34.5 & \cellcolor[RGB]{187,211,236}24.0 & \cellcolor[RGB]{191,214,237}25.9 & \cellcolor[RGB]{162,195,229}\underline{12.7} & \cellcolor[RGB]{158,193,228}\textbf{10.9} \\
    Gemma3\mbox{-}27B & 76.9 & \cellcolor[RGB]{227,236,246}41.8 & \cellcolor[RGB]{232,240,247}44.3 & \cellcolor[RGB]{218,231,243}37.9 & \cellcolor[RGB]{216,229,243}36.9 & \cellcolor[RGB]{178,206,233}19.9 & \cellcolor[RGB]{161,195,229}\underline{12.3} & \cellcolor[RGB]{192,215,237}26.4 & \cellcolor[RGB]{161,195,229}12.4 & \cellcolor[RGB]{160,194,229}\textbf{12.0} \\
    \midrule
    \multicolumn{11}{c}{\textit{AIME24}} \\
    \midrule
    Qwen2.5\mbox{-}7B & 12.6 & \cellcolor[RGB]{229,238,246}42.8 & \cellcolor[RGB]{233,240,247}44.7 & \cellcolor[RGB]{201,220,239}30.3 & \cellcolor[RGB]{230,238,247}43.3 & \cellcolor[RGB]{199,219,239}29.5 & \cellcolor[RGB]{176,204,233}19.0 & \cellcolor[RGB]{222,233,244}39.5 & \cellcolor[RGB]{166,198,230}\underline{14.7} & \cellcolor[RGB]{159,194,229}\textbf{11.5} \\
    Qwen2.5\mbox{-}14B & 13.8 & \cellcolor[RGB]{227,236,246}41.8 & \cellcolor[RGB]{232,239,247}44.0 & \cellcolor[RGB]{217,230,243}37.5 & \cellcolor[RGB]{228,237,246}42.3 & \cellcolor[RGB]{189,213,236}25.1 & \cellcolor[RGB]{168,199,231}\underline{15.6} & \cellcolor[RGB]{204,222,240}31.4 & \cellcolor[RGB]{162,196,229}\textbf{13.1} & \cellcolor[RGB]{171,201,232}16.9 \\
    Qwen2.5\mbox{-}32B & 16.9 & \cellcolor[RGB]{228,237,246}42.4 & \cellcolor[RGB]{233,240,247}44.4 & \cellcolor[RGB]{218,231,243}37.9 & \cellcolor[RGB]{231,239,247}43.5 & \cellcolor[RGB]{187,211,236}23.9 & \cellcolor[RGB]{171,201,231}\underline{16.7} & \cellcolor[RGB]{210,225,241}34.1 & \cellcolor[RGB]{157,192,228}\textbf{10.6} & \cellcolor[RGB]{179,206,234}20.4 \\
    Qwen3\mbox{-}8B & 27.7 & \cellcolor[RGB]{211,227,242}34.9 & \cellcolor[RGB]{221,233,244}39.3 & \cellcolor[RGB]{193,215,237}26.5 & \cellcolor[RGB]{215,229,243}36.4 & \cellcolor[RGB]{224,234,245}40.3 & \cellcolor[RGB]{229,238,246}42.9 & \cellcolor[RGB]{236,242,248}45.8 & \cellcolor[RGB]{165,197,230}\textbf{14.2} & \cellcolor[RGB]{167,199,230}\underline{14.9} \\
    Qwen3\mbox{-}14B & 27.1 & \cellcolor[RGB]{208,224,241}33.2 & \cellcolor[RGB]{219,231,244}38.1 & \cellcolor[RGB]{188,212,236}24.4 & \cellcolor[RGB]{213,228,242}35.7 & \cellcolor[RGB]{204,222,240}31.4 & \cellcolor[RGB]{169,200,231}\textbf{16.2} & \cellcolor[RGB]{232,240,247}44.1 & \cellcolor[RGB]{179,207,234}20.6 & \cellcolor[RGB]{174,203,232}\underline{18.1} \\
    Qwen3\mbox{-}32B & 28.3 & \cellcolor[RGB]{205,223,240}32.0 & \cellcolor[RGB]{217,230,243}37.4 & \cellcolor[RGB]{186,211,235}23.7 & --- & \cellcolor[RGB]{232,240,247}44.2 & \cellcolor[RGB]{217,230,243}37.6 & \cellcolor[RGB]{191,214,237}25.7 & \cellcolor[RGB]{182,208,234}\underline{21.6} & \cellcolor[RGB]{170,201,231}\textbf{16.4} \\
    Gemma3\mbox{-}12B & 23.9 & \cellcolor[RGB]{222,233,245}39.8 & \cellcolor[RGB]{229,238,246}42.8 & \cellcolor[RGB]{211,226,242}34.7 & \cellcolor[RGB]{211,226,242}34.6 & \cellcolor[RGB]{208,224,241}33.2 & \cellcolor[RGB]{204,222,240}31.8 & \cellcolor[RGB]{185,210,235}23.0 & \cellcolor[RGB]{155,191,227}\phantom{0}\textbf{9.7} & \cellcolor[RGB]{161,195,229}\underline{12.6} \\
    Gemma3\mbox{-}27B & 29.0 & \cellcolor[RGB]{225,235,245}40.7 & \cellcolor[RGB]{231,239,247}43.5 & \cellcolor[RGB]{214,228,242}36.2 & \cellcolor[RGB]{214,228,242}36.1 & \cellcolor[RGB]{200,219,239}29.6 & \cellcolor[RGB]{199,219,239}29.4 & \cellcolor[RGB]{180,207,234}\underline{21.1} & \cellcolor[RGB]{151,189,227}\phantom{0}\textbf{8.1} & \cellcolor[RGB]{188,212,236}24.3 \\
    \midrule
    \multicolumn{11}{c}{\textit{AIME25}} \\
    \midrule
    Qwen2.5\mbox{-}7B & \phantom{0}9.1 & \cellcolor[RGB]{228,237,246}42.4 & \cellcolor[RGB]{233,240,247}44.4 & \cellcolor[RGB]{203,221,240}31.0 & \cellcolor[RGB]{230,238,246}43.1 & \cellcolor[RGB]{193,215,237}26.6 & \cellcolor[RGB]{170,200,231}\underline{16.2} & \cellcolor[RGB]{217,230,243}37.3 & \cellcolor[RGB]{158,193,228}\textbf{11.2} & \cellcolor[RGB]{179,206,234}20.5 \\
    Qwen2.5\mbox{-}14B & 14.7 & \cellcolor[RGB]{228,237,246}42.4 & \cellcolor[RGB]{233,240,247}44.4 & \cellcolor[RGB]{218,231,243}37.8 & \cellcolor[RGB]{229,238,246}42.9 & \cellcolor[RGB]{195,216,238}27.5 & \cellcolor[RGB]{168,200,231}\underline{15.7} & \cellcolor[RGB]{204,222,240}31.6 & \cellcolor[RGB]{157,193,228}\textbf{10.8} & \cellcolor[RGB]{190,213,236}25.4 \\
    Qwen2.5\mbox{-}32B & 12.2 & \cellcolor[RGB]{229,238,246}42.7 & \cellcolor[RGB]{233,240,247}44.7 & \cellcolor[RGB]{219,231,244}38.3 & \cellcolor[RGB]{231,239,247}43.8 & \cellcolor[RGB]{202,221,239}30.6 & \cellcolor[RGB]{174,203,232}\underline{18.4} & \cellcolor[RGB]{207,224,241}33.0 & \cellcolor[RGB]{153,190,227}\phantom{0}\textbf{8.7} & \cellcolor[RGB]{197,217,238}28.3 \\
    Qwen3\mbox{-}8B & 22.5 & \cellcolor[RGB]{210,226,241}34.2 & \cellcolor[RGB]{220,232,244}38.8 & \cellcolor[RGB]{191,214,237}25.9 & \cellcolor[RGB]{214,228,242}36.0 & \cellcolor[RGB]{217,230,243}37.6 & \cellcolor[RGB]{230,239,247}43.4 & \cellcolor[RGB]{238,243,249}46.9 & \cellcolor[RGB]{161,195,229}\underline{12.3} & \cellcolor[RGB]{154,191,227}\phantom{0}\textbf{9.5} \\
    Qwen3\mbox{-}14B & 27.5 & \cellcolor[RGB]{206,223,240}32.6 & \cellcolor[RGB]{218,231,243}37.8 & \cellcolor[RGB]{187,211,236}24.0 & \cellcolor[RGB]{212,227,242}35.3 & \cellcolor[RGB]{202,221,239}30.8 & \cellcolor[RGB]{177,205,233}19.7 & \cellcolor[RGB]{252,252,252}53.0 & \cellcolor[RGB]{171,202,232}\underline{17.1} & \cellcolor[RGB]{145,185,225}\phantom{0}\textbf{5.3} \\
    Qwen3\mbox{-}32B & 23.7 & \cellcolor[RGB]{204,222,240}31.7 & \cellcolor[RGB]{217,230,243}37.2 & \cellcolor[RGB]{184,209,235}22.5 & --- & \cellcolor[RGB]{215,229,243}36.3 & \cellcolor[RGB]{218,231,243}37.8 & \cellcolor[RGB]{191,214,236}25.6 & \cellcolor[RGB]{176,204,233}\underline{19.1} & \cellcolor[RGB]{149,187,226}\phantom{0}\textbf{6.9} \\
    Gemma3\mbox{-}12B & 18.8 & \cellcolor[RGB]{223,234,245}40.3 & \cellcolor[RGB]{230,238,246}43.2 & \cellcolor[RGB]{213,227,242}35.5 & \cellcolor[RGB]{211,226,242}34.5 & \cellcolor[RGB]{205,223,240}32.1 & \cellcolor[RGB]{199,219,239}29.4 & \cellcolor[RGB]{169,200,231}\underline{15.9} & \cellcolor[RGB]{170,201,231}16.7 & \cellcolor[RGB]{151,189,226}\phantom{0}\textbf{7.8} \\
    Gemma3\mbox{-}27B & 25.3 & \cellcolor[RGB]{224,235,245}40.6 & \cellcolor[RGB]{230,239,247}43.4 & \cellcolor[RGB]{214,228,242}35.9 & \cellcolor[RGB]{213,227,242}35.4 & \cellcolor[RGB]{189,213,236}25.1 & \cellcolor[RGB]{184,209,235}22.7 & \cellcolor[RGB]{185,210,235}23.1 & \cellcolor[RGB]{168,199,231}\underline{15.6} & \cellcolor[RGB]{158,193,228}\textbf{11.1} \\
    \midrule
    \rowcolor{gray!20} \textbf{Average} & 48.0 & 39.3 & 42.4 & 32.7 & 39.0 & 30.8 & 23.6 & 35.0 & \underline{13.2} & \textbf{13.0} \\
    \bottomrule
  \end{tabular}}
  \sbox{\wbecerightbox}{%
  \begin{tabular}[t]{@{}wc{3.55em}@{}}
    \toprule
    \multirow{2}{*}{\centering\makecell[c]{\textbf{PRM}\\\textbf{(ref.)}}}\vphantom{\textbf{DTC}} \\
    \noalign{\vskip\aboverulesep\vskip\lightrulewidth\vskip\belowrulesep}
    \vphantom{$\mathrm{mean}$\dag} \\
    \midrule
    \vphantom{\textit{MATH-500}} \\
    \midrule
    \cellcolor[RGB]{155,191,228}\phantom{0}9.8 \\
    \cellcolor[RGB]{153,190,227}\phantom{0}8.8 \\
    \cellcolor[RGB]{152,189,227}\phantom{0}8.4 \\
    \cellcolor[RGB]{150,188,226}\phantom{0}7.4 \\
    \cellcolor[RGB]{147,186,225}\phantom{0}6.2 \\
    \cellcolor[RGB]{147,186,226}\phantom{0}6.3 \\
    \cellcolor[RGB]{149,188,226}\phantom{0}7.2 \\
    \cellcolor[RGB]{152,189,227}\phantom{0}8.4 \\
    \midrule
    \vphantom{\textit{AMC23}} \\
    \midrule
    \cellcolor[RGB]{147,186,225}\phantom{0}6.0 \\
    \cellcolor[RGB]{147,186,225}\phantom{0}6.0 \\
    \cellcolor[RGB]{153,190,227}\phantom{0}8.9 \\
    \cellcolor[RGB]{159,194,228}11.4 \\
    \cellcolor[RGB]{160,195,229}12.0 \\
    \cellcolor[RGB]{159,194,229}11.6 \\
    \cellcolor[RGB]{156,192,228}10.1 \\
    \cellcolor[RGB]{163,196,229}13.1 \\
    \midrule
    \vphantom{\textit{AIME24}} \\
    \midrule
    \cellcolor[RGB]{184,209,235}22.7 \\
    \cellcolor[RGB]{185,210,235}23.0 \\
    \cellcolor[RGB]{186,210,235}23.4 \\
    \cellcolor[RGB]{199,219,239}29.2 \\
    \cellcolor[RGB]{200,219,239}29.7 \\
    \cellcolor[RGB]{198,218,238}28.8 \\
    \cellcolor[RGB]{199,219,239}29.2 \\
    \cellcolor[RGB]{201,220,239}30.0 \\
    \midrule
    \vphantom{\textit{AIME25}} \\
    \midrule
    \cellcolor[RGB]{190,213,236}25.2 \\
    \cellcolor[RGB]{188,212,236}24.4 \\
    \cellcolor[RGB]{195,217,238}27.7 \\
    \cellcolor[RGB]{200,219,239}29.6 \\
    \cellcolor[RGB]{202,221,239}30.7 \\
    \cellcolor[RGB]{202,221,239}30.8 \\
    \cellcolor[RGB]{194,216,237}27.2 \\
    \cellcolor[RGB]{193,215,237}26.7 \\
    \midrule
    \rowcolor{gray!20} 18.1 \\
    \bottomrule
  \end{tabular}}
  \def\wbecevdash{%
    \smash{%
      \raisebox{-\dp\wbeceleftbox}{%
        \vbox to \dimexpr\ht\wbeceleftbox+\dp\wbeceleftbox\relax{%
          \offinterlineskip
          \xleaders\vbox{\hrule width 0.85pt height 3.8pt\vskip 1.15pt}\vfill
        }%
      }%
    }%
  }
  \setlength{\dimen0}{\dimexpr\wd\wbeceleftbox+0.70em+\wd\wbecerightbox\relax}
  \ifdim\dimen0>\textwidth
    \resizebox{\textwidth}{!}{%
    \begin{tabular}[t]{@{}c@{\hspace{0.35em}}c@{\hspace{0.35em}}c@{}}
    \usebox{\wbeceleftbox}&\wbecevdash&\usebox{\wbecerightbox}
    \end{tabular}}%
  \else
    \makebox[\textwidth][c]{%
    \begin{tabular}[t]{@{}c@{\hspace{0.35em}}c@{\hspace{0.35em}}c@{}}
    \usebox{\wbeceleftbox}&\wbecevdash&\usebox{\wbecerightbox}
    \end{tabular}}%
  \fi
\end{table}

\section{Experiments}
\label{sec:experiments}
\begin{figure}[!t]
  \centering
  \includegraphics[width=\linewidth]{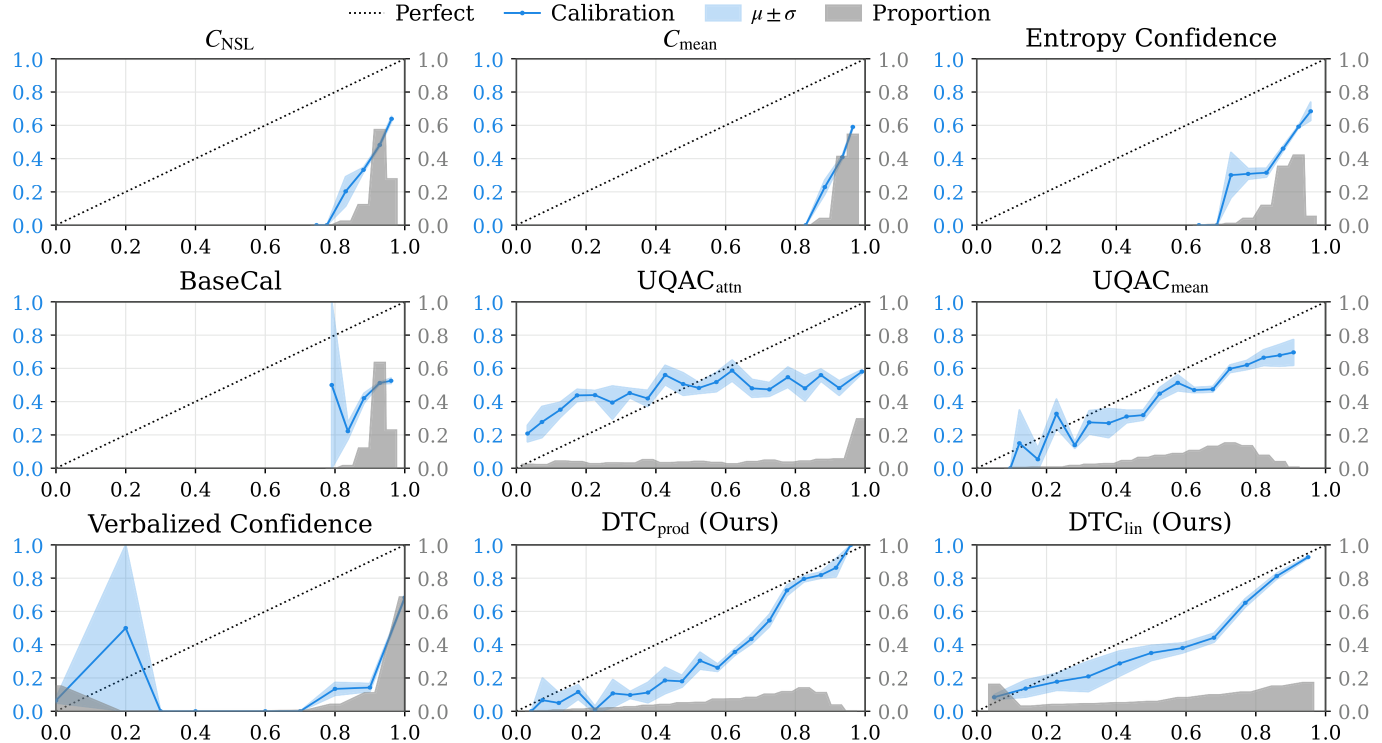}
  \caption{Calibration plots and probability histogram for Qwen2.5-14B on MATH-500.
  The $x$-axis shows mean confidence within 20 probability bins.
  The calibration curve (\textcolor[RGB]{30,136,229}{blue} line with $\mu\pm\sigma$) displays actual accuracy per bin, while the \textcolor{gray}{gray} shadow represents the probability proportion.}
  \label{fig:whitebox-calibration}
\end{figure}

\subsection{White-box Setting}
\label{sec:rq2}
\label{sec:whitebox}
\label{sec:whitebox-setup}


\paragraph{Datasets and Models.}
To evaluate \ours on tasks with long CoT path, we use four widely used mathematical benchmarks spanning a range of difficulty: MATH-500~\citep{Hendrycks-2021-MATH,Lightman-2023-Lets-Verify} and the contest sets AMC23, AIME24, and AIME25~\citep{Balunovic-2025-MathArena}.
We evaluate eight LLMs from three families: Qwen2.5-\{7,14,32\}B-Instruct~\citep{Qwen-2024-Qwen2.5}, Qwen3-\{8,14,32\}B~\citep{Yang-2025-Qwen3}, and Gemma3-\{12,27\}B-IT~\citep{Gemma-2025-Gemma3}.

\paragraph{Baselines.}
Given the limited prior work on calibrated reasoning uncertainty, we compare \ours against three classes of uncertainty estimators.
(1) \emph{Standard full-sequence confidence}: {\boldmath\(\CNSL\)} uses length-normalized sequence likelihood~(\Cref{eq:nsl}); {\boldmath\(\Cmean\)} uses mean token probability~(\Cref{sec:semantic-token-uq}); and \textbf{Entropy Confidence} that uses length-normalized predictive entropy confidence~\citep{Li-2025-UQAC}.
(2) \emph{Refinements of full-sequence confidence}: \textbf{BaseCal}~\citep{Tan-2026-BaseCal} averages response-token probabilities under a paired base model; {\boldmath\(\PGattn\)}~\citep{Li-2025-UQAC} further select attention-related tokens; and our variant {\boldmath\(\PGmean\)} which replaces each selected-token probability with the \(\Cmean\) before multiplying over the selected set.
BaseCal is omitted for Qwen3-32B because no public base checkpoint is available.
(3) \emph{Verbalized estimation}: \cite{Xiong-2024-Confidence-Elicitation} propose \textbf{Verbalized confidence} through additional prompts after the response elicits the model confidence.
We also include \textbf{PRM} (Skywork-o1-Open-PRM-Qwen-2.5-7B;~\citealp{He-2024-Skywork-o1}) as a supervised reference, averaging rewards over newline-delimited reasoning steps into path-level confidence.

\paragraph{Implementations.}
We use the officially recommended decoding parameters for each LLM, as detailed in~\Cref{app:js_token_impl}.
For each model family, we use its smallest model and 100 problems drawn at random from the MATH training set~\citep{Hendrycks-2021-MATH} to set that family's disagreement threshold \(\theta\).
We report \(\DTCprod\) and \(\DTClin\), with \(\theta=0.50\) for Qwen2.5, \(\theta=0.60\) for Qwen3, and \(\theta=0.85\) for Gemma3.
For \(\DTCprod\), token probabilities come from the generator \(\gG\)~(\Cref{eq:pprod}).
Unless noted, the auxiliary is the smaller same-family instruct model: Qwen2.5-1.5B-Instruct, Qwen3-1.7B, or Gemma3-4B-IT.
Other auxiliary sizes are examined in~\Cref{sec:analysis}.
Following~\citet{Li-2025-UQAC}, we evaluate ECE and AUROC by repeated subsampling, emphasizing ECE and the calibration plots.
AUROC is the area under the ROC curve and measures ranking. Tables and figures use AUC for AUROC.
Further details are in~\Cref{app:metrics}.

\paragraph{Main Results.}
\label{sec:whitebox-results}
\Cref{tab:whitebox-ece} and \Cref{tab:whitebox-auroc} (\Cref{app:full_tables}) report the ECE and AUROC results, respectively. \Cref{fig:whitebox-calibration} visualizes calibration curves on MATH-500 with Qwen2.5-14B. We find that: 

\textbf{Full-sequence confidence retains ranking information but remains overconfident.}
The three standard methods (\(\CNSL\), \(\Cmean\) and Entropy Confidence) achieve an high average AUROC of \(72.7\)--\(73.8\), yet their ECEs reach \(32.7\)--\(42.4\).
Their calibration curves lie below the diagonal, with scores concentrated near the high-confidence.

\textbf{Refinements improve calibration unevenly and weaken ranking.}
BaseCal offers limited calibration gains, whereas the UQAC variants reduce ECE more substantially.
UQAC variant \(\PGmean\) by ours achieves both the lowest average ECE (\(23.6\)) and highest AUROC (\(66.7\)) among these refinements, but its ranking still trails the standard methods.

\textbf{DTC improves both calibration and average ranking.}
\(\DTClin\) and \(\DTCprod\) reduce average ECE to \(13.0\) and \(13.2\), with calibration curves closer to the diagonal, while attaining AUROC \(75.3\) and \(80.8\).
Thus, count alone supports effective calibration; retaining path-mean probability further improves average ranking at similar ECE.
Relative to \(\PGmean\), \(\DTCprod\) lowers average ECE by \(10.4\), suggesting that selecting divergent tokens captures uncertainty more effectively and improves the calibration of \(\Cmean\).
In harder datasets, PRM has higher ECE, while \ours is better calibrated.

\subsection{Black-box Setting}
\label{sec:rq3}
\label{sec:blackbox}
\label{sec:blackbox-setup}

\paragraph{Datasets and Models.}
To better match a black-box evaluation setting, we use more challenging contest benchmarks than in the white-box experiments, namely AIME24, AIME25, HMMT25, and HMMT26~\citep{Balunovic-2025-MathArena}, and switch to stronger generators: Qwen3-4B-Instruct-2507, Qwen3-30B-A3B-Instruct-2507~\citep{Yang-2025-Qwen3}, and DeepSeek-V3.2~\citep{DeepSeek-2025-V3.2}.
We treat each generator as a black box, using its output trajectories without accessing its logits for scoring.
More details on datasets and models are shown in~\Cref{app:datasets_models}.

\begin{table}[!t]
  \centering
  \caption{Uncertainty quantification performance with DeepSeek-V3.2 in black-box settings.}
  \label{tab:blackbox-ece}
  \footnotesize
  \setlength{\tabcolsep}{1.0pt}%
  \renewcommand{\arraystretch}{1.05}%
  \begin{tabular}{@{}wl{7.4em}@{\hspace{2.0pt}}wc{2.55em}wc{2.80em}wc{2.80em}@{\hspace{2.0pt}}wc{2.55em}wc{2.80em}wc{2.80em}@{\hspace{2.0pt}}wc{2.55em}wc{2.80em}wc{2.80em}@{\hspace{2.0pt}}wc{2.55em}wc{2.80em}wc{2.80em}@{}}
    \toprule
    \multirow{2}{*}{\centering\textbf{Method}} & \multicolumn{3}{c}{\textbf{AIME24}} & \multicolumn{3}{c}{\textbf{AIME25}} & \multicolumn{3}{c}{\textbf{HMMT25}} & \multicolumn{3}{c}{\textbf{HMMT26}} \\
    \cmidrule(lr){2-4}\cmidrule(lr){5-7}\cmidrule(lr){8-10}\cmidrule(lr){11-13}
    & \textbf{Acc}$\uparrow$ & \textbf{ECE}$\downarrow$ & \textbf{AUC}$\uparrow$ & \textbf{Acc}$\uparrow$ & \textbf{ECE}$\downarrow$ & \textbf{AUC}$\uparrow$ & \textbf{Acc}$\uparrow$ & \textbf{ECE}$\downarrow$ & \textbf{AUC}$\uparrow$ & \textbf{Acc}$\uparrow$ & \textbf{ECE}$\downarrow$ & \textbf{AUC}$\uparrow$ \\
    \midrule
    Default CoT & 72.2 & -- & -- & 60.9 & -- & -- & 46.1 & -- & -- & 50.9 & -- & -- \\
    \textcolor[HTML]{0072B2}{$\rightarrow$}\,+ PRM & -- & 39.0 & \textbf{88.2} & -- & 39.5 & 80.4 & -- & 41.1 & 60.4 & -- & 38.1 & 75.1 \\
    \cellcolor[RGB]{219,231,244}\textcolor[HTML]{0072B2}{$\rightarrow$}\,+ $\DTClin$ & \cellcolor[RGB]{219,231,244}-- & \cellcolor[RGB]{219,231,244}\textbf{11.2} & \cellcolor[RGB]{219,231,244}74.6 & \cellcolor[RGB]{219,231,244}-- & \cellcolor[RGB]{219,231,244}\phantom{0}\textbf{9.4} & \cellcolor[RGB]{219,231,244}78.7 & \cellcolor[RGB]{219,231,244}-- & \cellcolor[RGB]{219,231,244}\textbf{12.7} & \cellcolor[RGB]{219,231,244}67.9 & \cellcolor[RGB]{219,231,244}-- & \cellcolor[RGB]{219,231,244}\textbf{12.5} & \cellcolor[RGB]{219,231,244}72.7 \\
    \cellcolor[RGB]{219,231,244}\textcolor[HTML]{0072B2}{$\rightarrow$}\,+ $\DTCprod$ & \cellcolor[RGB]{219,231,244}-- & \cellcolor[RGB]{219,231,244}24.6 & \cellcolor[RGB]{219,231,244}79.4 & \cellcolor[RGB]{219,231,244}-- & \cellcolor[RGB]{219,231,244}23.9 & \cellcolor[RGB]{219,231,244}\textbf{82.1} & \cellcolor[RGB]{219,231,244}-- & \cellcolor[RGB]{219,231,244}29.1 & \cellcolor[RGB]{219,231,244}\textbf{75.0} & \cellcolor[RGB]{219,231,244}-- & \cellcolor[RGB]{219,231,244}26.4 & \cellcolor[RGB]{219,231,244}\textbf{76.5} \\
    Verb. Conf. & 67.4 & 40.8 & \textbf{85.5} & 55.3 & 39.6 & \textbf{86.8} & 38.3 & 40.3 & \textbf{78.2} & 45.0 & 40.1 & \textbf{84.3} \\
    \cellcolor[RGB]{219,231,244}\textcolor[HTML]{0072B2}{$\rightarrow$}\,+ $\DTClin$ & \cellcolor[RGB]{219,231,244}-- & \cellcolor[RGB]{219,231,244}\textbf{13.5} & \cellcolor[RGB]{219,231,244}67.1 & \cellcolor[RGB]{219,231,244}-- & \cellcolor[RGB]{219,231,244}\phantom{0}\textbf{9.2} & \cellcolor[RGB]{219,231,244}73.5 & \cellcolor[RGB]{219,231,244}-- & \cellcolor[RGB]{219,231,244}\textbf{17.5} & \cellcolor[RGB]{219,231,244}59.5 & \cellcolor[RGB]{219,231,244}-- & \cellcolor[RGB]{219,231,244}\textbf{14.5} & \cellcolor[RGB]{219,231,244}68.9 \\
    \cellcolor[RGB]{219,231,244}\textcolor[HTML]{0072B2}{$\rightarrow$}\,+ $\DTCprod$ & \cellcolor[RGB]{219,231,244}-- & \cellcolor[RGB]{219,231,244}25.9 & \cellcolor[RGB]{219,231,244}73.5 & \cellcolor[RGB]{219,231,244}-- & \cellcolor[RGB]{219,231,244}25.4 & \cellcolor[RGB]{219,231,244}78.3 & \cellcolor[RGB]{219,231,244}-- & \cellcolor[RGB]{219,231,244}30.9 & \cellcolor[RGB]{219,231,244}67.0 & \cellcolor[RGB]{219,231,244}-- & \cellcolor[RGB]{219,231,244}26.1 & \cellcolor[RGB]{219,231,244}73.7 \\
    Verb. TopK & 64.6 & 32.0 & \textbf{88.4} & 52.8 & 32.6 & \textbf{85.7} & 27.1 & 31.6 & \textbf{79.3} & 35.1 & 32.0 & \textbf{83.6} \\
    \cellcolor[RGB]{219,231,244}\textcolor[HTML]{0072B2}{$\rightarrow$}\,+ $\DTClin$ & \cellcolor[RGB]{219,231,244}-- & \cellcolor[RGB]{219,231,244}\textbf{14.2} & \cellcolor[RGB]{219,231,244}66.0 & \cellcolor[RGB]{219,231,244}-- & \cellcolor[RGB]{219,231,244}\textbf{14.1} & \cellcolor[RGB]{219,231,244}66.9 & \cellcolor[RGB]{219,231,244}-- & \cellcolor[RGB]{219,231,244}\textbf{18.8} & \cellcolor[RGB]{219,231,244}56.7 & \cellcolor[RGB]{219,231,244}-- & \cellcolor[RGB]{219,231,244}\textbf{18.2} & \cellcolor[RGB]{219,231,244}60.9 \\
    \cellcolor[RGB]{219,231,244}\textcolor[HTML]{0072B2}{$\rightarrow$}\,+ $\DTCprod$ & \cellcolor[RGB]{219,231,244}-- & \cellcolor[RGB]{219,231,244}32.9 & \cellcolor[RGB]{219,231,244}71.4 & \cellcolor[RGB]{219,231,244}-- & \cellcolor[RGB]{219,231,244}31.4 & \cellcolor[RGB]{219,231,244}71.2 & \cellcolor[RGB]{219,231,244}-- & \cellcolor[RGB]{219,231,244}36.1 & \cellcolor[RGB]{219,231,244}61.5 & \cellcolor[RGB]{219,231,244}-- & \cellcolor[RGB]{219,231,244}31.1 & \cellcolor[RGB]{219,231,244}64.6 \\
    Verb. PD & 68.3 & 32.6 & \textbf{86.2} & 57.1 & 33.2 & \textbf{84.9} & 33.1 & 31.5 & \textbf{74.3} & 37.9 & 32.3 & \textbf{83.1} \\
    \cellcolor[RGB]{219,231,244}\textcolor[HTML]{0072B2}{$\rightarrow$}\,+ $\DTClin$ & \cellcolor[RGB]{219,231,244}-- & \cellcolor[RGB]{219,231,244}\textbf{15.1} & \cellcolor[RGB]{219,231,244}65.4 & \cellcolor[RGB]{219,231,244}-- & \cellcolor[RGB]{219,231,244}\textbf{13.5} & \cellcolor[RGB]{219,231,244}69.1 & \cellcolor[RGB]{219,231,244}-- & \cellcolor[RGB]{219,231,244}\textbf{19.2} & \cellcolor[RGB]{219,231,244}57.8 & \cellcolor[RGB]{219,231,244}-- & \cellcolor[RGB]{219,231,244}\textbf{14.6} & \cellcolor[RGB]{219,231,244}66.0 \\
    \cellcolor[RGB]{219,231,244}\textcolor[HTML]{0072B2}{$\rightarrow$}\,+ $\DTCprod$ & \cellcolor[RGB]{219,231,244}-- & \cellcolor[RGB]{219,231,244}35.3 & \cellcolor[RGB]{219,231,244}72.6 & \cellcolor[RGB]{219,231,244}-- & \cellcolor[RGB]{219,231,244}32.6 & \cellcolor[RGB]{219,231,244}74.1 & \cellcolor[RGB]{219,231,244}-- & \cellcolor[RGB]{219,231,244}36.2 & \cellcolor[RGB]{219,231,244}63.0 & \cellcolor[RGB]{219,231,244}-- & \cellcolor[RGB]{219,231,244}31.8 & \cellcolor[RGB]{219,231,244}69.3 \\
    \bottomrule
  \end{tabular}
\end{table}

\begin{figure}[!t]
  \centering
  \begin{subfigure}[t]{0.32\textwidth}
    \centering
    \includegraphics[width=\linewidth]{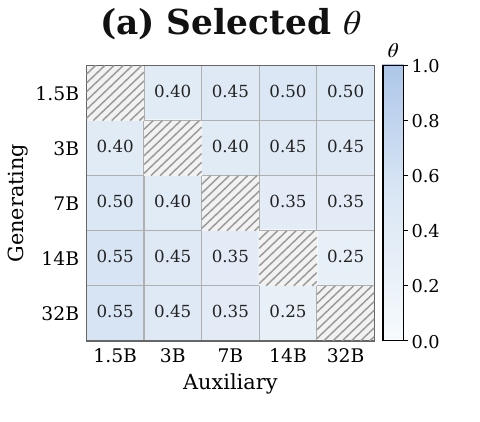}
  \end{subfigure}\hfill
  \begin{subfigure}[t]{0.32\textwidth}
    \centering
    \includegraphics[width=\linewidth]{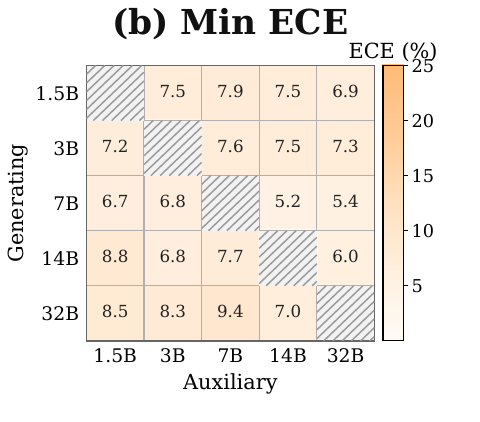}
  \end{subfigure}\hfill
  \begin{subfigure}[t]{0.32\textwidth}
    \centering
    \includegraphics[width=\linewidth]{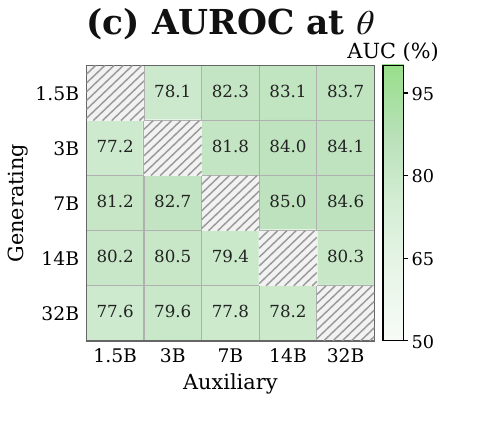}
  \end{subfigure}
  \caption{Sensitivity of $\DTClin$ to $\theta$ and auxiliary size, single-auxiliary Qwen2.5 on AMC23. We search $\theta$ in steps of $0.05$ and report the $\theta$ (a) with the lowest ECE (b) and its AUROC (c).}
  \label{fig:clin-theta}
\end{figure}

\paragraph{Baselines.}
Because generator token probabilities are unavailable, we use Verbalized estimation methods based on CoT prompts as the main confidence baselines~\citep{Wang-2026-BlackBox-UE}.
We consider three Verbalized estimation methods: \textbf{Verbalized Confidence} (Verb. Conf.), \textbf{Verbalized TopK} (Verb. TopK)~\citep{Tian-2023-Just-Ask}, and \textbf{Verbalized Probability Distribution} (Verb. PD)~\citep{Wang-2025-Verbalized-Distribution}.
As described in~\Cref{sec:related_work}, these Verbalized estimation methods change the CoT trajectory, so besides applying our method on \textbf{Default CoT}, we also apply it to the trajectories generated by these prompts for a fair comparison.

\paragraph{Evaluation and Implementation.}
Metrics, evaluation, and decoding follow the white-box setting; details appear in~\Cref{app:metrics,app:js_token_impl}.
All three generators use Qwen2.5-7B-Instruct as \(\gA'\) and Qwen2.5-1.5B-Instruct as \(\gA''\)~\citep{Qwen-2024-Qwen2.5}, with \(\theta{=}0.70\).

\paragraph{Main Results.}
\label{sec:blackbox-results}
\Cref{tab:blackbox-ece,tab:blackbox-qwen4b,tab:blackbox-qwen30b} report accuracy, ECE, and AUROC across three generators and four mathematics benchmarks, respectively. Blue arrows (\textcolor[HTML]{0072B2}{$\rightarrow$}) in the method column denote rescoring the same trajectories, leaving answers and accuracy unchanged.
PRM additionally scores Default CoT trajectories as a reference.
\textbf{Verbalized estimation is overconfident and sometimes reduces answer accuracy.}
Consistent with the accuracy and calibration limitations observed in mathematical reasoning settings by~\citet{Wang-2025-Verbalized-Distribution}, we find that Verbalized estimation can reduce answer accuracy while producing overconfident scores.
For DeepSeek-V3.2, the three Verbalized estimation methods reduce mean accuracy from \(57.5\) under Default CoT to \(44.9\)--\(51.5\), while their verbalized scores have mean ECE values of \(32.1\)--\(40.2\).

\textbf{DTC achieves low ECE on Default CoT and improves calibration over verbalized scores on the same trajectories.}
In contrast, our method assesses the reliability of Default CoT trajectories without changing the generated answers or reducing model accuracy.
On Default CoT, \(\DTClin\) achieves mean ECE values of \(11.5\)--\(12.7\) across the three generators.
When applied to trajectories generated under Verb.\ Conf., Verb.\ TopK, and Verb.\ PD, it also yields lower ECE than the original verbalized scores for every model--dataset pair.
On DeepSeek-V3.2, its mean ECE is \(13.7\)--\(16.3\), compared with \(32.1\)--\(40.2\) for the original verbalized scores.
\(\DTCprod\) achieves higher mean AUROC than \(\DTClin\), but also has higher mean ECE.
Despite these calibration gains, both estimators often have lower AUROC than the verbalized scores on trajectories from these three Verbalized estimation methods.

\Needspace{6\baselineskip}
\section{Analyses and Discussions}
\label{sec:analysis}
\begin{wrapfigure}[14]{r}{0.58\textwidth}
  \vspace{-1.6\baselineskip}%
  \centering
  \includegraphics[width=\linewidth]{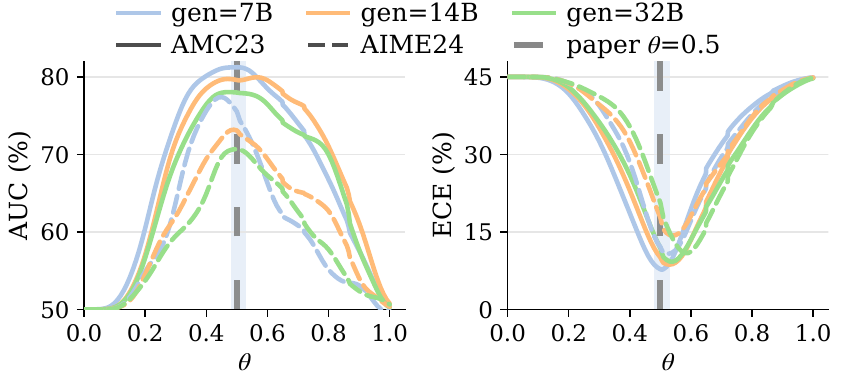}
  \caption{$\DTClin$ sensitivity to $\theta$ with auxiliary fixed at 1.5B
  and generating models $\in\{7,14,32\}$B (solid: AMC23; dashed: AIME24).
  The shaded band marks the white-box operating point $\theta{=}0.50$.}
  \label{fig:clin-theta-curves}
\end{wrapfigure}

\paragraph{Disagreement Threshold and auxiliary size.}
\label{sec:clin-theta}
We study how the choice of auxiliary model and disagreement threshold \(\theta\) affects \(\DTClin\). For each pair of generator and auxiliary, we search \(\theta\) in steps of \(0.05\) and report the \(\theta\) with the lowest ECE~(\Cref{fig:clin-theta}). The lowest ECE is similar across auxiliary sizes. At these values, a larger auxiliary usually has a higher AUROC. When the auxiliary is fixed at 1.5B, the \(\theta\) with the lowest ECE stays near \(0.5\)~(\Cref{fig:clin-theta-curves}). For efficiency and a simpler method, each model family uses one small auxiliary from the same family and one fixed \(\theta\).
The corresponding dual-auxiliary analyses are given in~\Cref{app:dual-aux}; the overall conclusions are similar to those in the single-auxiliary setting.

\vspace{-\baselineskip}
\begin{wraptable}[11]{r}{\dimexpr0.58\textwidth+30pt\relax}
  \vspace{-\baselineskip}
  \centering
  \caption{Verbalized vs.\ w/ DTC on DeepSeek-V3.2.}
  \label{tab:verbalized-dtc}
  \footnotesize
  \setlength{\tabcolsep}{1.0pt}%
  \renewcommand{\arraystretch}{1.05}%
  \begin{tabular}{@{}wl{4.6em}@{\hspace{2.0pt}}wc{2.80em}wc{2.80em}@{\hspace{2.0pt}}wc{2.80em}wc{2.80em}@{\hspace{2.0pt}}wc{2.80em}wc{2.80em}@{\hspace{2.0pt}}wc{2.80em}wc{2.80em}@{}}
    \toprule
    \multirow{2}{*}{\centering\textbf{Method}} & \multicolumn{2}{c}{\textbf{AIME24}} & \multicolumn{2}{c}{\textbf{AIME25}} & \multicolumn{2}{c}{\textbf{HMMT25}} & \multicolumn{2}{c}{\textbf{HMMT26}} \\
    \cmidrule(lr){2-3}\cmidrule(lr){4-5}\cmidrule(lr){6-7}\cmidrule(lr){8-9}
    & \textbf{ECE}$\downarrow$ & \textbf{AUC}$\uparrow$ & \textbf{ECE}$\downarrow$ & \textbf{AUC}$\uparrow$ & \textbf{ECE}$\downarrow$ & \textbf{AUC}$\uparrow$ & \textbf{ECE}$\downarrow$ & \textbf{AUC}$\uparrow$ \\
    \midrule
    Verb. Conf. & 40.8 & 85.5 & 39.6 & 86.8 & 40.3 & 78.2 & 40.1 & 84.3 \\
    \cellcolor[RGB]{219,231,244}w/ DTC & \cellcolor[RGB]{219,231,244}{\phantom{0}6.3} & \cellcolor[RGB]{219,231,244}{85.8} & \cellcolor[RGB]{219,231,244}{\phantom{0}3.7} & \cellcolor[RGB]{219,231,244}{87.7} & \cellcolor[RGB]{219,231,244}{10.5} & \cellcolor[RGB]{219,231,244}{77.8} & \cellcolor[RGB]{219,231,244}{\phantom{0}6.2} & \cellcolor[RGB]{219,231,244}{84.2} \\
    Verb. TopK & 32.0 & 88.4 & 32.6 & 85.7 & 31.6 & 79.3 & 32.0 & 83.6 \\
    \cellcolor[RGB]{219,231,244}w/ DTC & \cellcolor[RGB]{219,231,244}{17.0} & \cellcolor[RGB]{219,231,244}{90.3} & \cellcolor[RGB]{219,231,244}{14.6} & \cellcolor[RGB]{219,231,244}{87.3} & \cellcolor[RGB]{219,231,244}{18.8} & \cellcolor[RGB]{219,231,244}{79.1} & \cellcolor[RGB]{219,231,244}{15.7} & \cellcolor[RGB]{219,231,244}{83.6} \\
    Verb. PD & 32.6 & 86.2 & 33.2 & 84.9 & 31.5 & 74.3 & 32.3 & 83.1 \\
    \cellcolor[RGB]{219,231,244}w/ DTC & \cellcolor[RGB]{219,231,244}{10.2} & \cellcolor[RGB]{219,231,244}{87.7} & \cellcolor[RGB]{219,231,244}{10.9} & \cellcolor[RGB]{219,231,244}{86.9} & \cellcolor[RGB]{219,231,244}{19.2} & \cellcolor[RGB]{219,231,244}{76.0} & \cellcolor[RGB]{219,231,244}{13.9} & \cellcolor[RGB]{219,231,244}{84.0} \\
    \bottomrule
  \end{tabular}
\end{wraptable}

\paragraph{Combining verbalized scores with \ours.}
\label{sec:verbalized-dtc}
Verbalized estimation often yields overconfident scores~(high ECE) despite competitive ranking~(high AUROC).
We examine whether combining the divergent-token count with verbalized scores can reduce the overconfidence of Verbalized estimation, as \(\DTCprod\) does for \(\Cmean\).
We follow the way \(\DTCprod\) combines \(\Cmean\) with the uncertainty-informed count \(m\), and combine verbalized scores with DTC to get \(p_{\mathrm{verb}}^{m+k}\). The auxiliary and \(k\) follow the black-box \(\DTCprod\) setting. We report the result as w/ DTC in~\Cref{tab:verbalized-dtc,tab:verbalized-dtc-qwen30b,tab:verbalized-dtc-qwen4b}.
Across these tables, DTC lowers ECE in every case, often by more than 20 percentage points. AUROC changes little and sometimes improves.
On DeepSeek-V3.2, ECE decreases by 12.3 to 35.9 percentage points, and AUROC changes by \(-0.4\) to \(+2.0\) percentage points. Combining verbalized confidence with DTC thus gives a better tradeoff between ECE and AUROC.

\section{Conclusion}
\label{sec:conclusion}
We study the reliability of a reasoning path.
The UQAC pilot indicates that much of the calibration gain comes from how many tokens are selected.
\ours counts tokens on which two models disagree strongly about the next token.
\(\DTClin\) takes this count as the confidence.
\(\DTCprod\) recalibrates the trajectory's mean token probability with the same count, and a larger count gives a lower score.
On the math benchmarks, both improve calibration over sequence likelihood, verbalized confidence, and UQAC, whether the generator logits are available or not.
Applied to a verbalized score, the count also reduces overconfidence.



\section*{AI use statement}
We used generative AI tools to implement evaluation, plotting, and table-generation code, to assist with translation, and to draft and edit manuscript text, figures, and related-work notes.
We have not used generative AI tools to generate synthetic datasets, to formulate or prove mathematical claims, or to interpret experimental results, and the remaining required-disclosure tasks are not applicable.
Additionally, we used generative AI tools to search and format references and to improve readability.
We reviewed all AI-assisted text, citations, and code; numbers reported in the paper come from our evaluation pipeline and were checked by the authors.
We take responsibility for the final content of this work, including text, claims, and artifacts produced with the aid of generative AI.

\section*{Reproducibility statement}
Datasets, models, sampling hyperparameters, operating thresholds, and baseline protocols are specified in~\Cref{sec:experiments,app:experiment_details}.
Estimator definitions appear in~\Cref{sec:method}; additional ablations and black-box settings are in~\Cref{app:additional_analyses,app:threshold-sensitivity}.

\setlength{\bibsep}{1pt plus 1pt minus 0.5pt}
\bibliography{iclr2027_conference}
\bibliographystyle{iclr2027_conference}

\newpage
\appendix
\raggedbottom
\setlength{\intextsep}{6pt plus 1pt minus 2pt}
\makeatletter
\setlength{\@fptop}{0pt}
\setlength{\@fpsep}{8pt plus 2pt minus 2pt}
\setlength{\@fpbot}{0pt plus 1fil}
\makeatother

\section{Experimental Implementation Details}
\label{app:experiment_details}

This appendix records the experimental setup for~\Cref{sec:whitebox,sec:blackbox}: benchmarks and models, sampling, answer evaluation, and baselines.

\subsection{Datasets and Models}
\label{app:datasets_models}

We use four mathematical benchmarks in the white-box setting and four in the black-box setting. White-box experiments use MATH-500, AMC23, AIME24, and AIME25; black-box experiments use AIME24, AIME25, HMMT25, and HMMT26. \Cref{tab:app-datasets} lists the generators, auxiliary models, and the number of samples per question.

\begin{table}[H]
  \centering
  \caption{Benchmarks, models, and sample counts per question. The three black-box generators share one auxiliary pair.}
  \label{tab:app-datasets}
  \footnotesize
  \setlength{\tabcolsep}{8pt}
  \renewcommand{\arraystretch}{1.08}
  \resizebox{\textwidth}{!}{%
  \begin{tabular}{@{}c ll@{}}
    \toprule
    & \multicolumn{1}{c}{\textbf{White-box}} & \multicolumn{1}{c}{\textbf{Black-box}} \\
    \midrule
    \textbf{Benchmarks}
      & \begin{tabular}{@{}l@{}}
          MATH-500~\citep{Lightman-2023-Lets-Verify} \\
          AMC23~\citep{Balunovic-2025-MathArena} \\
          AIME24~\citep{Balunovic-2025-MathArena} \\
          AIME25~\citep{Balunovic-2025-MathArena}
        \end{tabular}
      & \begin{tabular}{@{}l@{}}
          AIME24~\citep{Balunovic-2025-MathArena} \\
          AIME25~\citep{Balunovic-2025-MathArena} \\
          HMMT25~\citep{Balunovic-2025-MathArena} \\
          HMMT26~\citep{Balunovic-2025-MathArena}
        \end{tabular} \\
    \midrule
    \textbf{Generators}
      & \begin{tabular}{@{}l@{}}
          Qwen2.5-7B-Instruct~\citep{Qwen-2024-Qwen2.5} \\
          Qwen2.5-14B-Instruct~\citep{Qwen-2024-Qwen2.5} \\
          Qwen2.5-32B-Instruct~\citep{Qwen-2024-Qwen2.5} \\
          Qwen3-8B~\citep{Yang-2025-Qwen3} \\
          Qwen3-14B~\citep{Yang-2025-Qwen3} \\
          Qwen3-32B~\citep{Yang-2025-Qwen3} \\
          Gemma3-12B-IT~\citep{Gemma-2025-Gemma3} \\
          Gemma3-27B-IT~\citep{Gemma-2025-Gemma3}
        \end{tabular}
      & \begin{tabular}{@{}l@{}}
          DeepSeek-V3.2~\citep{DeepSeek-2025-V3.2} \\
          Qwen3-30B-A3B-Instruct-2507~\citep{Yang-2025-Qwen3} \\
          Qwen3-4B-Instruct-2507~\citep{Yang-2025-Qwen3}
        \end{tabular} \\
    \midrule
    \textbf{Auxiliaries}
      & \begin{tabular}{@{}l@{}}
          Qwen2.5-1.5B-Instruct~\citep{Qwen-2024-Qwen2.5} \\
          Qwen3-1.7B~\citep{Yang-2025-Qwen3} \\
          Gemma3-4B-IT~\citep{Gemma-2025-Gemma3}
        \end{tabular}
      & \begin{tabular}{@{}l@{}}
          Qwen2.5-7B-Instruct~\citep{Qwen-2024-Qwen2.5} \\
          Qwen2.5-1.5B-Instruct~\citep{Qwen-2024-Qwen2.5}
        \end{tabular} \\
    \midrule
    \begin{tabular}{@{}c@{}}
      \textbf{Sample number} \\
      \textbf{per question}
    \end{tabular}
      & \begin{tabular}{@{}l@{}}
          $8$ on MATH-500 \\
          $64$ on AMC23 \\
          $64$ on AIME24 \\
          $64$ on AIME25
        \end{tabular}
      & \begin{tabular}{@{}l@{}}
          $32$ for DeepSeek-V3.2 \\
          $64$ for Qwen3-30B-A3B-Instruct-2507 \\
          $64$ for Qwen3-4B-Instruct-2507
        \end{tabular} \\
    \bottomrule
  \end{tabular}}
\end{table}

White-box generators are Qwen2.5-\{7B,14B,32B\}-Instruct, Qwen3-\{8B,14B,32B\}, and Gemma3-\{12B,27B\}-IT, paired with the smaller same-family auxiliaries Qwen2.5-1.5B-Instruct, Qwen3-1.7B, and Gemma3-4B-IT. Black-box generators are DeepSeek-V3.2, Qwen3-30B-A3B-Instruct-2507, and Qwen3-4B-Instruct-2507. In the main black-box setting, all three are scored with the external pair Qwen2.5-7B-Instruct and Qwen2.5-1.5B-Instruct.

\subsection{Sampling and Prompting}
\label{app:sampling}
\label{app:js_token_impl}

We use a zero-shot system prompt and pass the problem as the user message:

\begin{prompt}[title={Reasoning Prompt}]
Please reason step by step, and put your final answer within \texttt{\textbackslash boxed\{\}}.
\end{prompt}

For the black-box protocols in~\Cref{sec:blackbox}, we add each protocol's confidence instruction at generation time. We then score these trajectories and do not generate new ones.

We decode each generator with its own recommended setting. We set the maximum sequence length to $8192$ tokens for white-box generators and to $16384$ tokens for black-box generators. Under vLLM, we use temperature $0.6$ and top-$p=0.95$ for Qwen2.5-\{7B,14B,32B\}-Instruct, Qwen3-\{8B,14B,32B\}, and Gemma3-\{12B,27B\}-IT. For Qwen3-4B-Instruct-2507 and Qwen3-30B-A3B-Instruct-2507, we use temperature $0.7$, top-$p=0.8$, top-$k=20$, and presence penalty $1.0$. We sample DeepSeek-V3.2 from its API at temperature $1.0$ and top-$p=0.95$. We draw $8$ trajectories per MATH-500 problem and $64$ per contest problem for vLLM generators, and $32$ per problem for DeepSeek-V3.2.

\subsection{Answer Extraction and Evaluation}
\label{app:metrics}

Following~\citet{Li-2025-UQAC}, we separate answer generation from uncertainty quantification.
Once a response is generated, we extract the final answer and check whether it is correct.
If no answer is extracted, we exclude that instance from the UQ evaluation.
We then compute confidence on the same trajectory with teacher forcing. This step does not generate a new answer.
Accuracy counts every sampled response, including truncated ones.

Following~\citet{Li-2025-UQAC}, we then subsample to balance the number of correct and incorrect predictions.
If there are fewer correct predictions than incorrect ones, we randomly sample a matching number of incorrect predictions.
If both groups exceed $500$ instances, we randomly select $500$ from each group.
This step does not change the confidence scores, but it does change AUROC and ECE.
We repeat the subsampling five times and report the mean and standard deviation.
Accuracy is left unchanged.

Following~\citet{Li-2025-UQAC}, we take ECE (use $20$ equal-width bins in our experiments) as the primary metric and AUROC as secondary, because AUROC only shows whether correct answers rank above incorrect ones.
It does not show whether a confidence value can be read as a probability of being correct.
If the confidences are $0.9$, $0.5$, and $0.1$, with the first two answers correct and the third incorrect, AUROC is $1$.
Replacing these scores with $9\times 10^{-3}$, $8\times 10^{-10}$, and $7.99\times 10^{-10}$ keeps the ranking, so AUROC remains $1$, even though the scores are not interpretable as probabilities.
A high AUROC does not mean that confidence matches the probability of correctness.
AUROC can still be $1$ when the highest confidence among $10{,}000$ answers is only $1\times 10^{-3}$, so the scores can still misrepresent how certain the model is.




\subsection{Baselines}
\label{app:baselines}

\subsubsection{White-box baselines}


\paragraph{Length-normalized sequence likelihood.}
The product of token probabilities shrinks as the trajectory grows, so we take its geometric mean~(\Cref{eq:nsl};~\citealp{Malinin-2021-Predictive-Entropy}). With \(p_t=P_{\gG}(\traj_t\mid x,\traj_{<t})\) and length \(T\),
\begin{equation}
  \CNSL
  \;=\;
  \Bigl(\prod_{t=1}^{T} p_t\Bigr)^{1/T}.
\end{equation}

\paragraph{Mean token probability.}
The arithmetic mean also removes the length effect, and a single low-probability token moves it less than a product~\citep{Orgad-2025-Know-More}:
\begin{equation}
  \Cmean
  \;=\;
  \frac{1}{T}\sum_{t=1}^{T} p_t.
\end{equation}

\paragraph{Entropy confidence.}
Predictive entropy aggregates the next-token entropy over the trajectory and has no fixed range~\citep{Kuhn-2023-Semantic-Uncertainty}.
Length-normalized predictive entropy divides that sum by the trajectory length, so outputs of different lengths remain comparable~\citep{Malinin-2021-Predictive-Entropy}.
We report the confidence as one minus this normalized entropy:
\begin{equation}
  H(\traj)
  \;=\;
  -\sum_{t=1}^{T}
  \sum_{v\in\sV}
  P_{\gG}(v\mid x,\traj_{<t})\log P_{\gG}(v\mid x,\traj_{<t}),
  \qquad
  1-\bar H(\traj)
  \;=\;
  1-\frac{H(\traj)}{T}.
\end{equation}


\paragraph{BaseCal.}
BaseCal scores a post-trained generator with its paired base model, which is often better calibrated~\citep{Tan-2026-BaseCal}.
BaseCal-ReEval passes the same trajectory through the base model and averages the probabilities of the generated tokens:
\begin{equation}
  C_{\mathrm{ReEval}}(\traj, x)
  \;=\;
  \frac{1}{T}\sum_{t=1}^{T}
  P_{\mathrm{base}}(\traj_t\mid x,\traj_{<t}).
\end{equation}
BaseCal-Proj instead trains a one-layer linear map from the generator's final-layer hidden states into the base model's hidden states, then reads token probabilities from the base output layer.
On our out-of-domain data this variant calibrates poorly, so we report only BaseCal-ReEval as BaseCal.

\paragraph{UQAC.}
Many tokens in a long trajectory say little about the answer. UQAC backtracks from the answer along attention and multiplies only the selected token probabilities~\citep{Li-2025-UQAC}. With selected set \(\mathcal{S}\),
\begin{equation}
  \PGattn
  \;=\;
  \prod_{t\in\mathcal{S}} p_t.
  \label{eq:app-uqac-attn}
\end{equation}
We replace every selected probability with the trajectory mean \(\Cmean\), so the score moves mainly with the number of selected tokens:
\begin{equation}
  \PGmean
  \;=\;
  \Cmean^{\,|\mathcal{S}|}.
\end{equation}

\paragraph{Verbalized estimation.}
We obtain the score by an additional prompt after the response~\citep{Xiong-2024-Confidence-Elicitation}.
The prompt asks the model to rate its previous answer from 0 to 10 and place the integer in \texttt{\textbackslash boxed\{\}}.
We divide that integer by 10, which gives a confidence in \([0,1]\).

\paragraph{Process-reward reference.}
We also report Skywork-o1-Open-PRM-Qwen-2.5-7B as a supervised reference~\citep{He-2024-Skywork-o1}. We split the response at newlines. For each step we apply a sigmoid to the flagged token value and then average the step rewards. PRM is left out of the best and second-best marks in the white-box table.


\begin{figure}[H]
  \centering
  \begin{prompt}[title={Verbalized confidence}, boxsep=2pt, top=1pt, bottom=1pt, before upper={\setlength{\parskip}{0pt}}]
    \textbf{Question:} \texttt{[Question]}

    Reason step-by-step to formulate your final answer.
    Your answer must be a mathematical value or short exact answer (e.g., an integer, fraction, radical, coordinate, or simplified expression), not a full sentence.
    If the question specifies a final report form (e.g., \(m{+}n\) or \(m{-}n\)), use that form.
    Then, reason about the confidence in your answer.
    Conclude by providing a JSON object that states the final answer and your estimated confidence in it:
    {\ttfamily
    \{\\
    "final\_answer": "Your final answer",\\
    "confidence": "0-1"\\
    \}\par}
  \end{prompt}
  \caption{Prompt for Verbalized confidence.
  \texttt{[Question]} is replaced by the problem.}
  \label{fig:app-prompt-verb-conf}
\end{figure}

\begin{figure}[H]
  \centering
  \begin{prompt}[title={Verbalized TopK}, boxsep=2pt, top=1pt, bottom=1pt, before upper={\setlength{\parskip}{0pt}}]
    \textbf{Question:} \texttt{[Question]}

    Reason step-by-step to formulate 2 best guesses and probability that each is correct.
    Each answer must be a mathematical value or short exact answer (e.g., an integer, fraction, radical, coordinate, or simplified expression), not a full sentence.
    If the question specifies a final report form (e.g., \(m{+}n\) or \(m{-}n\)), use that form.
    Your final output must be a JSON array:
    {\ttfamily
    [\newline
    \{\newline
    "candidate": "first most likely answer",\newline
    "confidence": "0-1"\newline
    \},\newline
    \{\newline
    "candidate": "second most likely answer",\newline
    "confidence": "0-1"\newline
    \}\newline
    ]\par}
  \end{prompt}
  \caption{Prompt for Verbalized TopK with \(k{=}2\).
  \texttt{[Question]} is replaced by the problem.}
  \label{fig:app-prompt-verb-topk}
\end{figure}

\begin{figure}[H]
  \centering
  \begin{prompt}[title={Verbalized probability distribution}, boxsep=2pt, top=1pt, bottom=1pt, before upper={\setlength{\parskip}{0pt}}]
    \textbf{Question:} \texttt{[Question]}

    Reason step-by-step to formulate your answer.
    You may propose multiple possible answers (fewer than five).
    Each answer must be a mathematical value or short exact answer (e.g., an integer, fraction, radical, coordinate, or simplified expression), not a full sentence.
    If the question specifies a final report form (e.g., \(m{+}n\) or \(m{-}n\)), use that form.
    Always include ``None of the above'' as a possible answer.
    Reason about the confidence in each possible answer.
    Your final output must be a JSON array where the confidence scores form a probability distribution (they must sum to 1.0):
    {\ttfamily
    [\newline
    \{\newline
    "candidate": "Candidate 1",\newline
    "confidence": "0-1"\newline
    \},\newline
    \{\newline
    "candidate": "Candidate 2",\newline
    "confidence": "0-1"\newline
    \},\newline
    ...\newline
    \{\newline
    "candidate": "None of the above",\newline
    "confidence": "0-1"\newline
    \}\newline
    ]\par}
  \end{prompt}
  \caption{Prompt for Verbalized probability distribution.
  \texttt{[Question]} is replaced by the problem.}
  \label{fig:app-prompt-verb-dist}
\end{figure}

\subsubsection{Black-box baselines}

\paragraph{Verbalized confidence.}
We ask the model to write the answer and a confidence score in the same generation~\citep{Tian-2023-Just-Ask}.
The prompt is shown in~\Cref{fig:app-prompt-verb-conf}.
A single stated number is easy to read, but the model can focus on the answer it already prefers.

\paragraph{Verbalized TopK.}
We ask the model to list \(k{=}2\) candidate answers, each with a probability, and we take the probability of the highest-scoring candidate~\citep{Tian-2023-Just-Ask}.
The prompt is shown in~\Cref{fig:app-prompt-verb-topk}.

\paragraph{Verbalized probability distribution.}
We ask the model to assign probabilities to candidate answers that sum to \(1\).
On open-ended questions the list includes ``None of the above'', which holds the remaining low-probability answers.
The prompt is shown in~\Cref{fig:app-prompt-verb-dist}.
The reported confidence is the probability of the selected answer, so that mass has to be shared with the other candidates.

\section{Complete Results}
\label{app:more_results}

The tables below add the results omitted from the main text. Accuracy uses every sampled response. ECE and AUROC use the class-balanced sets in~\Cref{app:metrics}. An arrow marks a row that only rescores the same trajectories, so that row has no accuracy of its own. A dash means the metric is missing or does not apply.

\subsection{Full White-box Results}
\label{app:full_tables}

The main text reports white-box ECE~(\Cref{tab:whitebox-ece}). \Cref{tab:whitebox-auroc} is the AUROC table for the same single-auxiliary setup, with the same generators, benchmarks, and baselines. PRM is a supervised reference and is not part of the unsupervised best and second-best comparison.

\subsection{Full Black-box Results}
\label{app:blackbox_full}

\Cref{tab:blackbox-qwen4b,tab:blackbox-qwen30b} are the Qwen3-4B and Qwen3-30B-A3B results. Each table gives accuracy for the original generation protocol, plus ECE and AUROC on AIME24, AIME25, HMMT25, and HMMT26, with averages. DTC and PRM rescore the frozen Default CoT trajectories. Within a confidence protocol, that protocol and its DTC rows use the same trajectories.

\subsection{Verbalized Confidence with DTC}
\label{app:verbalized_dtc}

\Cref{tab:verbalized-dtc-qwen30b,tab:verbalized-dtc-qwen4b} repeat the verbalized-confidence experiment on Qwen3-30B-A3B and Qwen3-4B. For each protocol, the DTC adjustment uses the black-box \(\DTCprod\) setup and the same frozen trajectory as the unadjusted verbalized score, so accuracy does not change.

\begin{table}[H]
  \centering
  \footnotesize
  \setlength{\tabcolsep}{2.0pt}
  \renewcommand{\arraystretch}{1.05}
  \caption{Uncertainty quantification performance (AUROC, $\uparrow$ higher is better) in white-box settings.
  In each row, the best result is in \textbf{bold} and the second-best one
  is \underline{underlined} (excluding PRM).
  \dag : UQAC variant by ours.}
  \label{tab:whitebox-auroc}
  \ifcsname wbaurocleftbox\endcsname\else\newsavebox{\wbaurocleftbox}\fi
  \ifcsname wbaurocrightbox\endcsname\else\newsavebox{\wbaurocrightbox}\fi
  \sbox{\wbaurocleftbox}{%
  \begin{tabular}[t]{@{}wl{5.6em}wc{2.3em}wc{3.55em}wc{3.55em}wc{3.55em}wc{3.55em}wc{3.55em}wc{3.55em}wc{3.55em}wc{3.55em}wc{3.55em}@{}}
    \toprule
    \multirow{2}{*}{\makecell[l]{\textbf{Reasoning} \\ \textbf{Models}}} & \multirow{2}{*}{\centering\textbf{Acc}} & \multirow{2}{*}{\centering{\boldmath$\CNSL$}} & \multirow{2}{*}{\centering{\boldmath$\Cmean$}} & \multirow{2}{*}{\centering\makecell[c]{\textbf{Entropy}\\\textbf{Conf.}}} & \multirow{2}{*}{\centering\textbf{BaseCal}} & \multicolumn{2}{c}{\textbf{UQAC}} & \multirow{2}{*}{\centering\textbf{Verb.}} & \multicolumn{2}{c}{\textbf{DTC (Ours)}} \\
    \cmidrule(lr){7-8}\cmidrule(lr){10-11}
    & & & & & & \textbf{attn} & \textbf{mean}\textsuperscript{\dag} & & \textbf{prod} & \textbf{lin} \\
    \midrule
    \multicolumn{11}{c}{\textit{MATH-500}} \\
    \midrule
    Qwen2.5\mbox{-}7B & 76.1 & \cellcolor[RGB]{195,216,238}67.0 & \cellcolor[RGB]{196,217,238}66.5 & \cellcolor[RGB]{206,223,240}61.4 & \cellcolor[RGB]{212,227,242}58.5 & \cellcolor[RGB]{211,226,242}59.2 & \cellcolor[RGB]{205,222,240}62.0 & \cellcolor[RGB]{179,206,234}74.7 & \cellcolor[RGB]{167,199,231}\underline{80.4} & \cellcolor[RGB]{160,195,229}\textbf{83.8} \\
    Qwen2.5\mbox{-}14B & 80.0 & \cellcolor[RGB]{197,218,238}65.7 & \cellcolor[RGB]{199,219,239}65.0 & \cellcolor[RGB]{199,219,239}65.0 & \cellcolor[RGB]{217,230,243}56.0 & \cellcolor[RGB]{216,229,243}56.6 & \cellcolor[RGB]{199,219,239}64.9 & \cellcolor[RGB]{179,206,234}74.8 & \cellcolor[RGB]{164,197,230}\underline{81.9} & \cellcolor[RGB]{158,193,228}\textbf{85.0} \\
    Qwen2.5\mbox{-}32B & 82.4 & \cellcolor[RGB]{195,216,238}66.7 & \cellcolor[RGB]{197,218,238}65.8 & \cellcolor[RGB]{198,218,238}65.5 & \cellcolor[RGB]{228,237,246}50.6 & \cellcolor[RGB]{210,225,241}59.7 & \cellcolor[RGB]{199,219,239}65.0 & \cellcolor[RGB]{160,195,229}\underline{83.8} & \cellcolor[RGB]{163,196,229}82.7 & \cellcolor[RGB]{156,192,228}\textbf{85.8} \\
    Qwen3\mbox{-}8B & 83.9 & \cellcolor[RGB]{178,206,233}\underline{75.0} & \cellcolor[RGB]{182,208,234}73.4 & \cellcolor[RGB]{184,209,235}72.4 & \cellcolor[RGB]{225,235,245}52.3 & \cellcolor[RGB]{211,226,242}59.1 & \cellcolor[RGB]{206,223,240}61.6 & \cellcolor[RGB]{213,228,242}57.9 & \cellcolor[RGB]{169,200,231}\textbf{79.3} & \cellcolor[RGB]{180,207,234}73.9 \\
    Qwen3\mbox{-}14B & 86.5 & \cellcolor[RGB]{180,207,234}74.3 & \cellcolor[RGB]{183,209,235}72.8 & \cellcolor[RGB]{184,210,235}72.0 & \cellcolor[RGB]{223,234,245}53.2 & \cellcolor[RGB]{196,217,238}66.1 & \cellcolor[RGB]{177,205,233}\underline{75.6} & \cellcolor[RGB]{205,223,240}61.8 & \cellcolor[RGB]{169,200,231}\textbf{79.4} & \cellcolor[RGB]{177,205,233}75.6 \\
    Qwen3\mbox{-}32B & 83.7 & \cellcolor[RGB]{181,208,234}73.6 & \cellcolor[RGB]{185,210,235}71.6 & \cellcolor[RGB]{187,211,236}70.9 & --- & \cellcolor[RGB]{211,227,242}58.8 & \cellcolor[RGB]{203,221,240}63.1 & \cellcolor[RGB]{202,221,239}63.3 & \cellcolor[RGB]{167,199,231}\textbf{80.6} & \cellcolor[RGB]{170,201,231}\underline{78.9} \\
    Gemma3\mbox{-}12B & 84.9 & \cellcolor[RGB]{205,223,240}61.9 & \cellcolor[RGB]{208,225,241}60.3 & \cellcolor[RGB]{203,222,240}62.7 & \cellcolor[RGB]{248,249,251}40.9 & \cellcolor[RGB]{205,223,240}61.9 & \cellcolor[RGB]{213,227,242}58.1 & \cellcolor[RGB]{160,194,229}\textbf{84.1} & \cellcolor[RGB]{164,197,230}81.9 & \cellcolor[RGB]{160,194,229}\underline{84.0} \\
    Gemma3\mbox{-}27B & 89.2 & \cellcolor[RGB]{202,220,239}63.6 & \cellcolor[RGB]{205,223,240}61.8 & \cellcolor[RGB]{202,221,239}63.3 & \cellcolor[RGB]{252,252,252}38.9 & \cellcolor[RGB]{175,204,233}76.5 & \cellcolor[RGB]{165,197,230}81.7 & \cellcolor[RGB]{153,190,227}\underline{87.2} & \cellcolor[RGB]{157,192,228}85.5 & \cellcolor[RGB]{152,189,227}\textbf{87.9} \\
    \midrule
    \multicolumn{11}{c}{\textit{AMC23}} \\
    \midrule
    Qwen2.5\mbox{-}7B & 53.6 & \cellcolor[RGB]{184,209,235}72.4 & \cellcolor[RGB]{185,210,235}71.8 & \cellcolor[RGB]{193,215,237}67.9 & \cellcolor[RGB]{195,217,238}66.6 & \cellcolor[RGB]{201,220,239}63.9 & \cellcolor[RGB]{192,215,237}68.1 & \cellcolor[RGB]{194,216,237}67.1 & \cellcolor[RGB]{166,198,230}\textbf{81.3} & \cellcolor[RGB]{166,198,230}\underline{81.2} \\
    Qwen2.5\mbox{-}14B & 61.2 & \cellcolor[RGB]{195,216,238}66.9 & \cellcolor[RGB]{196,217,238}66.4 & \cellcolor[RGB]{193,215,237}67.7 & \cellcolor[RGB]{204,222,240}62.3 & \cellcolor[RGB]{215,229,243}57.1 & \cellcolor[RGB]{192,214,237}68.5 & \cellcolor[RGB]{182,208,234}73.0 & \cellcolor[RGB]{170,201,231}\underline{79.2} & \cellcolor[RGB]{168,199,231}\textbf{80.3} \\
    Qwen2.5\mbox{-}32B & 66.4 & \cellcolor[RGB]{194,216,237}67.1 & \cellcolor[RGB]{196,217,238}66.3 & \cellcolor[RGB]{189,212,236}69.8 & \cellcolor[RGB]{207,224,241}60.8 & \cellcolor[RGB]{204,222,240}62.2 & \cellcolor[RGB]{191,214,237}68.6 & \cellcolor[RGB]{181,208,234}73.7 & \cellcolor[RGB]{173,202,232}\underline{77.6} & \cellcolor[RGB]{172,202,232}\textbf{78.0} \\
    Qwen3\mbox{-}8B & 68.6 & \cellcolor[RGB]{181,207,234}73.8 & \cellcolor[RGB]{184,210,235}72.0 & \cellcolor[RGB]{185,210,235}71.6 & \cellcolor[RGB]{217,230,243}56.2 & \cellcolor[RGB]{248,249,251}41.1 & \cellcolor[RGB]{234,241,247}47.7 & \cellcolor[RGB]{213,228,242}57.8 & \cellcolor[RGB]{171,201,232}\textbf{78.7} & \cellcolor[RGB]{176,205,233}\underline{76.0} \\
    Qwen3\mbox{-}14B & 74.1 & \cellcolor[RGB]{181,208,234}73.5 & \cellcolor[RGB]{185,210,235}71.5 & \cellcolor[RGB]{183,209,235}72.6 & \cellcolor[RGB]{216,229,243}56.6 & \cellcolor[RGB]{196,217,238}66.1 & \cellcolor[RGB]{183,209,234}72.9 & \cellcolor[RGB]{191,214,237}68.8 & \cellcolor[RGB]{170,201,231}\textbf{79.2} & \cellcolor[RGB]{178,205,233}\underline{75.3} \\
    Qwen3\mbox{-}32B & 67.5 & \cellcolor[RGB]{178,205,233}75.3 & \cellcolor[RGB]{182,208,234}73.4 & \cellcolor[RGB]{180,207,234}74.4 & --- & \cellcolor[RGB]{219,231,244}55.1 & \cellcolor[RGB]{215,229,243}57.2 & \cellcolor[RGB]{193,215,237}67.7 & \cellcolor[RGB]{167,199,231}\textbf{80.3} & \cellcolor[RGB]{171,201,232}\underline{78.6} \\
    Gemma3\mbox{-}12B & 66.8 & \cellcolor[RGB]{189,213,236}69.7 & \cellcolor[RGB]{191,214,237}68.6 & \cellcolor[RGB]{188,212,236}70.0 & \cellcolor[RGB]{221,232,244}54.2 & \cellcolor[RGB]{206,223,241}61.2 & \cellcolor[RGB]{205,223,240}61.9 & \cellcolor[RGB]{171,201,231}\underline{78.7} & \cellcolor[RGB]{168,200,231}\textbf{79.9} & \cellcolor[RGB]{171,201,231}78.7 \\
    Gemma3\mbox{-}27B & 76.9 & \cellcolor[RGB]{191,214,237}68.5 & \cellcolor[RGB]{194,216,237}67.1 & \cellcolor[RGB]{192,214,237}68.5 & \cellcolor[RGB]{234,241,247}47.8 & \cellcolor[RGB]{183,209,235}72.7 & \cellcolor[RGB]{176,204,233}76.4 & \cellcolor[RGB]{157,192,228}\textbf{85.6} & \cellcolor[RGB]{163,196,229}\underline{82.6} & \cellcolor[RGB]{167,199,231}80.4 \\
    \midrule
    \multicolumn{11}{c}{\textit{AIME24}} \\
    \midrule
    Qwen2.5\mbox{-}7B & 12.6 & \cellcolor[RGB]{170,201,231}\underline{79.0} & \cellcolor[RGB]{172,202,232}78.0 & \cellcolor[RGB]{180,207,234}74.4 & \cellcolor[RGB]{178,206,233}75.2 & \cellcolor[RGB]{219,231,244}55.1 & \cellcolor[RGB]{206,223,240}61.5 & \cellcolor[RGB]{170,201,231}79.0 & \cellcolor[RGB]{159,194,229}\textbf{84.4} & \cellcolor[RGB]{177,205,233}75.6 \\
    Qwen2.5\mbox{-}14B & 13.8 & \cellcolor[RGB]{186,211,235}71.3 & \cellcolor[RGB]{186,211,235}71.0 & \cellcolor[RGB]{178,206,233}75.2 & \cellcolor[RGB]{198,218,238}65.5 & \cellcolor[RGB]{206,223,240}61.4 & \cellcolor[RGB]{188,212,236}70.1 & \cellcolor[RGB]{173,202,232}\textbf{77.8} & \cellcolor[RGB]{175,204,233}\underline{76.5} & \cellcolor[RGB]{180,207,234}74.2 \\
    Qwen2.5\mbox{-}32B & 16.9 & \cellcolor[RGB]{177,205,233}75.6 & \cellcolor[RGB]{178,206,233}75.2 & \cellcolor[RGB]{170,200,231}79.3 & \cellcolor[RGB]{184,209,235}72.2 & \cellcolor[RGB]{194,215,237}67.5 & \cellcolor[RGB]{181,208,234}73.5 & \cellcolor[RGB]{160,194,229}\textbf{84.1} & \cellcolor[RGB]{168,199,231}\underline{80.0} & \cellcolor[RGB]{188,212,236}70.4 \\
    Qwen3\mbox{-}8B & 27.7 & \cellcolor[RGB]{186,211,235}71.3 & \cellcolor[RGB]{189,212,236}69.9 & \cellcolor[RGB]{184,210,235}\underline{72.1} & \cellcolor[RGB]{212,227,242}58.4 & \cellcolor[RGB]{222,233,244}53.5 & \cellcolor[RGB]{216,230,243}56.4 & \cellcolor[RGB]{198,218,238}65.5 & \cellcolor[RGB]{180,207,234}\textbf{74.1} & \cellcolor[RGB]{197,218,238}65.7 \\
    Qwen3\mbox{-}14B & 27.1 & \cellcolor[RGB]{187,211,236}70.8 & \cellcolor[RGB]{190,213,236}69.4 & \cellcolor[RGB]{187,211,236}70.6 & \cellcolor[RGB]{206,223,240}61.4 & \cellcolor[RGB]{199,219,239}64.9 & \cellcolor[RGB]{183,209,235}\underline{72.6} & \cellcolor[RGB]{201,220,239}63.9 & \cellcolor[RGB]{180,207,234}\textbf{74.3} & \cellcolor[RGB]{195,216,238}67.0 \\
    Qwen3\mbox{-}32B & 28.3 & \cellcolor[RGB]{186,211,235}71.2 & \cellcolor[RGB]{189,212,236}69.9 & \cellcolor[RGB]{182,208,234}73.4 & --- & \cellcolor[RGB]{225,235,245}52.1 & \cellcolor[RGB]{224,235,245}52.5 & \cellcolor[RGB]{179,206,234}\underline{74.7} & \cellcolor[RGB]{176,204,233}\textbf{76.2} & \cellcolor[RGB]{189,213,236}69.6 \\
    Gemma3\mbox{-}12B & 23.9 & \cellcolor[RGB]{166,198,230}\underline{81.0} & \cellcolor[RGB]{168,199,231}80.1 & \cellcolor[RGB]{168,200,231}79.8 & \cellcolor[RGB]{186,211,235}71.1 & \cellcolor[RGB]{199,219,239}64.8 & \cellcolor[RGB]{207,224,241}60.9 & \cellcolor[RGB]{179,206,234}74.8 & \cellcolor[RGB]{164,197,230}\textbf{81.8} & \cellcolor[RGB]{196,217,238}66.4 \\
    Gemma3\mbox{-}27B & 29.0 & \cellcolor[RGB]{171,201,232}78.5 & \cellcolor[RGB]{172,202,232}78.1 & \cellcolor[RGB]{172,202,232}78.3 & \cellcolor[RGB]{192,215,237}68.2 & \cellcolor[RGB]{196,217,238}66.4 & \cellcolor[RGB]{199,219,239}65.1 & \cellcolor[RGB]{152,189,227}\textbf{87.8} & \cellcolor[RGB]{168,199,231}\underline{80.2} & \cellcolor[RGB]{201,220,239}64.0 \\
    \midrule
    \multicolumn{11}{c}{\textit{AIME25}} \\
    \midrule
    Qwen2.5\mbox{-}7B & \phantom{0}9.1 & \cellcolor[RGB]{169,200,231}\textbf{79.3} & \cellcolor[RGB]{170,201,231}\underline{78.9} & \cellcolor[RGB]{173,202,232}77.8 & \cellcolor[RGB]{171,201,231}78.7 & \cellcolor[RGB]{193,215,237}67.7 & \cellcolor[RGB]{186,211,235}71.1 & \cellcolor[RGB]{189,213,236}69.6 & \cellcolor[RGB]{175,204,233}76.5 & \cellcolor[RGB]{200,220,239}64.3 \\
    Qwen2.5\mbox{-}14B & 14.7 & \cellcolor[RGB]{178,206,233}75.1 & \cellcolor[RGB]{179,206,234}74.7 & \cellcolor[RGB]{176,205,233}\textbf{75.9} & \cellcolor[RGB]{185,210,235}71.8 & \cellcolor[RGB]{192,214,237}68.5 & \cellcolor[RGB]{177,205,233}\underline{75.5} & \cellcolor[RGB]{180,207,234}74.4 & \cellcolor[RGB]{177,205,233}\textbf{75.9} & \cellcolor[RGB]{204,222,240}62.2 \\
    Qwen2.5\mbox{-}32B & 12.2 & \cellcolor[RGB]{170,201,231}79.0 & \cellcolor[RGB]{170,201,231}\underline{79.1} & \cellcolor[RGB]{169,200,231}\textbf{79.4} & \cellcolor[RGB]{177,205,233}75.4 & \cellcolor[RGB]{224,235,245}52.4 & \cellcolor[RGB]{173,203,232}77.5 & \cellcolor[RGB]{184,210,235}72.1 & \cellcolor[RGB]{175,204,233}76.7 & \cellcolor[RGB]{209,225,241}60.1 \\
    Qwen3\mbox{-}8B & 22.5 & \cellcolor[RGB]{170,201,231}\underline{78.9} & \cellcolor[RGB]{173,203,232}77.4 & \cellcolor[RGB]{174,203,232}77.2 & \cellcolor[RGB]{193,215,237}67.9 & \cellcolor[RGB]{219,231,244}55.1 & \cellcolor[RGB]{217,230,243}56.2 & \cellcolor[RGB]{202,221,239}63.3 & \cellcolor[RGB]{163,196,230}\textbf{82.4} & \cellcolor[RGB]{187,211,236}70.9 \\
    Qwen3\mbox{-}14B & 27.5 & \cellcolor[RGB]{163,196,230}82.5 & \cellcolor[RGB]{165,198,230}81.3 & \cellcolor[RGB]{162,196,229}82.8 & \cellcolor[RGB]{181,207,234}73.8 & \cellcolor[RGB]{193,215,237}67.7 & \cellcolor[RGB]{160,194,229}\underline{84.1} & \cellcolor[RGB]{225,235,245}52.1 & \cellcolor[RGB]{149,188,226}\textbf{89.1} & \cellcolor[RGB]{164,197,230}81.8 \\
    Qwen3\mbox{-}32B & 23.7 & \cellcolor[RGB]{167,199,231}80.3 & \cellcolor[RGB]{171,201,231}78.8 & \cellcolor[RGB]{165,198,230}\underline{81.3} & --- & \cellcolor[RGB]{206,223,241}61.2 & \cellcolor[RGB]{205,223,240}61.7 & \cellcolor[RGB]{179,206,234}74.8 & \cellcolor[RGB]{154,191,227}\textbf{86.8} & \cellcolor[RGB]{168,200,231}79.9 \\
    Gemma3\mbox{-}12B & 18.8 & \cellcolor[RGB]{158,193,228}84.9 & \cellcolor[RGB]{160,194,229}84.1 & \cellcolor[RGB]{159,194,229}84.4 & \cellcolor[RGB]{191,214,236}69.0 & \cellcolor[RGB]{199,219,239}64.8 & \cellcolor[RGB]{195,216,238}66.7 & \cellcolor[RGB]{156,192,228}\underline{85.7} & \cellcolor[RGB]{150,188,226}\textbf{88.9} & \cellcolor[RGB]{176,204,233}76.3 \\
    Gemma3\mbox{-}27B & 25.3 & \cellcolor[RGB]{154,191,227}86.9 & \cellcolor[RGB]{156,192,228}85.8 & \cellcolor[RGB]{155,191,228}86.4 & \cellcolor[RGB]{194,216,237}67.2 & \cellcolor[RGB]{187,211,236}70.9 & \cellcolor[RGB]{179,206,234}74.8 & \cellcolor[RGB]{145,185,225}\textbf{91.3} & \cellcolor[RGB]{148,187,226}\underline{89.8} & \cellcolor[RGB]{170,201,231}79.0 \\
    \midrule
    \rowcolor{gray!20} \textbf{Average} & 48.0 & 73.8 & 72.7 & 73.2 & 61.8 & 61.6 & 66.7 & 73.5 & \textbf{80.8} & \underline{75.3} \\
    \bottomrule
  \end{tabular}}
  \sbox{\wbaurocrightbox}{%
  \begin{tabular}[t]{@{}wc{3.55em}@{}}
    \toprule
    \multirow{2}{*}{\centering\makecell[c]{\textbf{PRM}\\\textbf{(ref.)}}}\vphantom{\textbf{DTC}} \\
    \noalign{\vskip\aboverulesep\vskip\lightrulewidth\vskip\belowrulesep}
    \vphantom{$\mathrm{mean}$\dag} \\
    \midrule
    \vphantom{\textit{MATH-500}} \\
    \midrule
    \cellcolor[RGB]{145,185,225}95.4 \\
    \cellcolor[RGB]{145,185,225}94.3 \\
    \cellcolor[RGB]{145,185,225}94.4 \\
    \cellcolor[RGB]{146,186,225}90.9 \\
    \cellcolor[RGB]{148,187,226}89.8 \\
    \cellcolor[RGB]{145,185,225}91.3 \\
    \cellcolor[RGB]{145,185,225}93.6 \\
    \cellcolor[RGB]{145,185,225}93.2 \\
    \midrule
    \vphantom{\textit{AMC23}} \\
    \midrule
    \cellcolor[RGB]{146,186,225}90.8 \\
    \cellcolor[RGB]{147,186,225}90.5 \\
    \cellcolor[RGB]{153,190,227}87.2 \\
    \cellcolor[RGB]{146,186,225}90.6 \\
    \cellcolor[RGB]{149,187,226}89.5 \\
    \cellcolor[RGB]{150,188,226}89.0 \\
    \cellcolor[RGB]{155,191,227}86.6 \\
    \cellcolor[RGB]{146,185,225}91.1 \\
    \midrule
    \vphantom{\textit{AIME24}} \\
    \midrule
    \cellcolor[RGB]{145,185,225}96.2 \\
    \cellcolor[RGB]{148,187,226}89.7 \\
    \cellcolor[RGB]{148,187,226}89.6 \\
    \cellcolor[RGB]{150,188,226}89.0 \\
    \cellcolor[RGB]{150,188,226}88.6 \\
    \cellcolor[RGB]{145,185,225}91.7 \\
    \cellcolor[RGB]{145,185,225}93.8 \\
    \cellcolor[RGB]{146,186,225}90.8 \\
    \midrule
    \vphantom{\textit{AIME25}} \\
    \midrule
    \cellcolor[RGB]{151,188,226}88.6 \\
    \cellcolor[RGB]{151,189,227}88.3 \\
    \cellcolor[RGB]{160,195,229}83.8 \\
    \cellcolor[RGB]{153,190,227}87.5 \\
    \cellcolor[RGB]{153,190,227}87.5 \\
    \cellcolor[RGB]{152,189,227}88.1 \\
    \cellcolor[RGB]{148,187,226}89.6 \\
    \cellcolor[RGB]{145,185,225}91.1 \\
    \midrule
    \rowcolor{gray!20} 90.4 \\
    \bottomrule
  \end{tabular}}
  \def\wbaurocvdash{%
    \smash{%
      \raisebox{-\dp\wbaurocleftbox}{%
        \vbox to \dimexpr\ht\wbaurocleftbox+\dp\wbaurocleftbox\relax{%
          \offinterlineskip
          \xleaders\vbox{\hrule width 0.85pt height 3.8pt\vskip 1.15pt}\vfill
        }%
      }%
    }%
  }
  \setlength{\dimen0}{\dimexpr\wd\wbaurocleftbox+0.70em+\wd\wbaurocrightbox\relax}
  \ifdim\dimen0>\textwidth
    \resizebox{\textwidth}{!}{%
    \begin{tabular}[t]{@{}c@{\hspace{0.35em}}c@{\hspace{0.35em}}c@{}}
    \usebox{\wbaurocleftbox}&\wbaurocvdash&\usebox{\wbaurocrightbox}
    \end{tabular}}%
  \else
    \makebox[\textwidth][c]{%
    \begin{tabular}[t]{@{}c@{\hspace{0.35em}}c@{\hspace{0.35em}}c@{}}
    \usebox{\wbaurocleftbox}&\wbaurocvdash&\usebox{\wbaurocrightbox}
    \end{tabular}}%
  \fi
\end{table}

\begin{table}[H]
  \centering
  \caption{Uncertainty quantification performance with Qwen3-4B in black-box settings.}
  \label{tab:blackbox-qwen4b}
  \footnotesize
  \setlength{\tabcolsep}{1.0pt}%
  \renewcommand{\arraystretch}{1.05}%
  \begin{tabular}{@{}wl{7.4em}@{\hspace{2.0pt}}wc{2.55em}wc{2.80em}wc{2.80em}@{\hspace{2.0pt}}wc{2.55em}wc{2.80em}wc{2.80em}@{\hspace{2.0pt}}wc{2.55em}wc{2.80em}wc{2.80em}@{\hspace{2.0pt}}wc{2.55em}wc{2.80em}wc{2.80em}@{}}
    \toprule
    \multirow{2}{*}{\centering\textbf{Method}} & \multicolumn{3}{c}{\textbf{AIME24}} & \multicolumn{3}{c}{\textbf{AIME25}} & \multicolumn{3}{c}{\textbf{HMMT25}} & \multicolumn{3}{c}{\textbf{HMMT26}} \\
    \cmidrule(lr){2-4}\cmidrule(lr){5-7}\cmidrule(lr){8-10}\cmidrule(lr){11-13}
    & \textbf{Acc}$\uparrow$ & \textbf{ECE}$\downarrow$ & \textbf{AUC}$\uparrow$ & \textbf{Acc}$\uparrow$ & \textbf{ECE}$\downarrow$ & \textbf{AUC}$\uparrow$ & \textbf{Acc}$\uparrow$ & \textbf{ECE}$\downarrow$ & \textbf{AUC}$\uparrow$ & \textbf{Acc}$\uparrow$ & \textbf{ECE}$\downarrow$ & \textbf{AUC}$\uparrow$ \\
    \midrule
    Default CoT & 61.4 & -- & -- & 45.6 & -- & -- & 30.3 & -- & -- & 34.7 & -- & -- \\
    \textcolor[HTML]{0072B2}{$\rightarrow$}\,+ PRM & -- & 36.0 & \textbf{82.9} & -- & 36.6 & 80.3 & -- & 37.8 & 53.3 & -- & 33.4 & 78.0 \\
    \cellcolor[RGB]{219,231,244}\textcolor[HTML]{0072B2}{$\rightarrow$}\,+ $\DTClin$ & \cellcolor[RGB]{219,231,244}-- & \cellcolor[RGB]{219,231,244}\textbf{11.3} & \cellcolor[RGB]{219,231,244}72.9 & \cellcolor[RGB]{219,231,244}-- & \cellcolor[RGB]{219,231,244}\phantom{0}\textbf{9.0} & \cellcolor[RGB]{219,231,244}81.5 & \cellcolor[RGB]{219,231,244}-- & \cellcolor[RGB]{219,231,244}\textbf{13.1} & \cellcolor[RGB]{219,231,244}69.6 & \cellcolor[RGB]{219,231,244}-- & \cellcolor[RGB]{219,231,244}\textbf{13.0} & \cellcolor[RGB]{219,231,244}74.6 \\
    \cellcolor[RGB]{219,231,244}\textcolor[HTML]{0072B2}{$\rightarrow$}\,+ $\DTCprod$ & \cellcolor[RGB]{219,231,244}-- & \cellcolor[RGB]{219,231,244}16.3 & \cellcolor[RGB]{219,231,244}75.8 & \cellcolor[RGB]{219,231,244}-- & \cellcolor[RGB]{219,231,244}10.7 & \cellcolor[RGB]{219,231,244}\textbf{84.8} & \cellcolor[RGB]{219,231,244}-- & \cellcolor[RGB]{219,231,244}15.4 & \cellcolor[RGB]{219,231,244}\textbf{78.0} & \cellcolor[RGB]{219,231,244}-- & \cellcolor[RGB]{219,231,244}14.3 & \cellcolor[RGB]{219,231,244}\textbf{79.0} \\
    Verb. Conf. & 61.8 & 45.6 & \textbf{80.8} & 46.6 & 45.9 & \textbf{85.6} & 27.3 & 45.6 & \textbf{81.9} & 33.0 & 45.1 & 79.8 \\
    \cellcolor[RGB]{219,231,244}\textcolor[HTML]{0072B2}{$\rightarrow$}\,+ $\DTClin$ & \cellcolor[RGB]{219,231,244}-- & \cellcolor[RGB]{219,231,244}\textbf{11.9} & \cellcolor[RGB]{219,231,244}71.6 & \cellcolor[RGB]{219,231,244}-- & \cellcolor[RGB]{219,231,244}\phantom{0}\textbf{9.9} & \cellcolor[RGB]{219,231,244}80.2 & \cellcolor[RGB]{219,231,244}-- & \cellcolor[RGB]{219,231,244}16.5 & \cellcolor[RGB]{219,231,244}66.5 & \cellcolor[RGB]{219,231,244}-- & \cellcolor[RGB]{219,231,244}12.9 & \cellcolor[RGB]{219,231,244}77.1 \\
    \cellcolor[RGB]{219,231,244}\textcolor[HTML]{0072B2}{$\rightarrow$}\,+ $\DTCprod$ & \cellcolor[RGB]{219,231,244}-- & \cellcolor[RGB]{219,231,244}17.4 & \cellcolor[RGB]{219,231,244}74.7 & \cellcolor[RGB]{219,231,244}-- & \cellcolor[RGB]{219,231,244}14.3 & \cellcolor[RGB]{219,231,244}84.4 & \cellcolor[RGB]{219,231,244}-- & \cellcolor[RGB]{219,231,244}\textbf{16.1} & \cellcolor[RGB]{219,231,244}73.7 & \cellcolor[RGB]{219,231,244}-- & \cellcolor[RGB]{219,231,244}\textbf{11.6} & \cellcolor[RGB]{219,231,244}\textbf{81.3} \\
    Verb. TopK & 52.4 & 47.0 & \textbf{68.3} & 42.0 & 47.0 & 75.3 & 26.4 & 46.8 & \textbf{74.0} & 33.3 & 45.5 & \textbf{77.3} \\
    \cellcolor[RGB]{219,231,244}\textcolor[HTML]{0072B2}{$\rightarrow$}\,+ $\DTClin$ & \cellcolor[RGB]{219,231,244}-- & \cellcolor[RGB]{219,231,244}\textbf{19.3} & \cellcolor[RGB]{219,231,244}60.6 & \cellcolor[RGB]{219,231,244}-- & \cellcolor[RGB]{219,231,244}\textbf{16.2} & \cellcolor[RGB]{219,231,244}74.0 & \cellcolor[RGB]{219,231,244}-- & \cellcolor[RGB]{219,231,244}\textbf{22.1} & \cellcolor[RGB]{219,231,244}54.2 & \cellcolor[RGB]{219,231,244}-- & \cellcolor[RGB]{219,231,244}\phantom{0}\textbf{9.9} & \cellcolor[RGB]{219,231,244}70.9 \\
    \cellcolor[RGB]{219,231,244}\textcolor[HTML]{0072B2}{$\rightarrow$}\,+ $\DTCprod$ & \cellcolor[RGB]{219,231,244}-- & \cellcolor[RGB]{219,231,244}30.1 & \cellcolor[RGB]{219,231,244}61.8 & \cellcolor[RGB]{219,231,244}-- & \cellcolor[RGB]{219,231,244}31.4 & \cellcolor[RGB]{219,231,244}\textbf{79.0} & \cellcolor[RGB]{219,231,244}-- & \cellcolor[RGB]{219,231,244}30.2 & \cellcolor[RGB]{219,231,244}62.7 & \cellcolor[RGB]{219,231,244}-- & \cellcolor[RGB]{219,231,244}25.0 & \cellcolor[RGB]{219,231,244}75.1 \\
    Verb. PD & 59.3 & 46.9 & 69.9 & 46.9 & 48.1 & 73.7 & 26.7 & 43.9 & \textbf{75.5} & 32.5 & 42.9 & 70.9 \\
    \cellcolor[RGB]{219,231,244}\textcolor[HTML]{0072B2}{$\rightarrow$}\,+ $\DTClin$ & \cellcolor[RGB]{219,231,244}-- & \cellcolor[RGB]{219,231,244}\textbf{14.4} & \cellcolor[RGB]{219,231,244}68.7 & \cellcolor[RGB]{219,231,244}-- & \cellcolor[RGB]{219,231,244}\phantom{0}\textbf{9.4} & \cellcolor[RGB]{219,231,244}76.3 & \cellcolor[RGB]{219,231,244}-- & \cellcolor[RGB]{219,231,244}\textbf{14.8} & \cellcolor[RGB]{219,231,244}65.6 & \cellcolor[RGB]{219,231,244}-- & \cellcolor[RGB]{219,231,244}\phantom{0}\textbf{7.2} & \cellcolor[RGB]{219,231,244}77.0 \\
    \cellcolor[RGB]{219,231,244}\textcolor[HTML]{0072B2}{$\rightarrow$}\,+ $\DTCprod$ & \cellcolor[RGB]{219,231,244}-- & \cellcolor[RGB]{219,231,244}25.7 & \cellcolor[RGB]{219,231,244}\textbf{72.6} & \cellcolor[RGB]{219,231,244}-- & \cellcolor[RGB]{219,231,244}22.4 & \cellcolor[RGB]{219,231,244}\textbf{82.1} & \cellcolor[RGB]{219,231,244}-- & \cellcolor[RGB]{219,231,244}22.7 & \cellcolor[RGB]{219,231,244}\textbf{75.5} & \cellcolor[RGB]{219,231,244}-- & \cellcolor[RGB]{219,231,244}20.1 & \cellcolor[RGB]{219,231,244}\textbf{82.2} \\
    \bottomrule
  \end{tabular}
\end{table}

\begin{table}[H]
  \centering
  \caption{Uncertainty quantification performance with Qwen3-30B-A3B in black-box settings.}
  \label{tab:blackbox-qwen30b}
  \footnotesize
  \setlength{\tabcolsep}{1.0pt}%
  \renewcommand{\arraystretch}{1.05}%
  \begin{tabular}{@{}wl{7.4em}@{\hspace{2.0pt}}wc{2.55em}wc{2.80em}wc{2.80em}@{\hspace{2.0pt}}wc{2.55em}wc{2.80em}wc{2.80em}@{\hspace{2.0pt}}wc{2.55em}wc{2.80em}wc{2.80em}@{\hspace{2.0pt}}wc{2.55em}wc{2.80em}wc{2.80em}@{}}
    \toprule
    \multirow{2}{*}{\centering\textbf{Method}} & \multicolumn{3}{c}{\textbf{AIME24}} & \multicolumn{3}{c}{\textbf{AIME25}} & \multicolumn{3}{c}{\textbf{HMMT25}} & \multicolumn{3}{c}{\textbf{HMMT26}} \\
    \cmidrule(lr){2-4}\cmidrule(lr){5-7}\cmidrule(lr){8-10}\cmidrule(lr){11-13}
    & \textbf{Acc}$\uparrow$ & \textbf{ECE}$\downarrow$ & \textbf{AUC}$\uparrow$ & \textbf{Acc}$\uparrow$ & \textbf{ECE}$\downarrow$ & \textbf{AUC}$\uparrow$ & \textbf{Acc}$\uparrow$ & \textbf{ECE}$\downarrow$ & \textbf{AUC}$\uparrow$ & \textbf{Acc}$\uparrow$ & \textbf{ECE}$\downarrow$ & \textbf{AUC}$\uparrow$ \\
    \midrule
    Default CoT & 73.4 & -- & -- & 59.8 & -- & -- & 42.4 & -- & -- & 43.5 & -- & -- \\
    \textcolor[HTML]{0072B2}{$\rightarrow$}\,+ PRM & -- & 37.5 & \textbf{86.0} & -- & 37.8 & 78.4 & -- & 38.7 & 53.6 & -- & 34.6 & 77.4 \\
    \cellcolor[RGB]{219,231,244}\textcolor[HTML]{0072B2}{$\rightarrow$}\,+ $\DTClin$ & \cellcolor[RGB]{219,231,244}-- & \cellcolor[RGB]{219,231,244}\textbf{15.9} & \cellcolor[RGB]{219,231,244}70.2 & \cellcolor[RGB]{219,231,244}-- & \cellcolor[RGB]{219,231,244}\textbf{12.3} & \cellcolor[RGB]{219,231,244}75.8 & \cellcolor[RGB]{219,231,244}-- & \cellcolor[RGB]{219,231,244}\textbf{13.1} & \cellcolor[RGB]{219,231,244}72.1 & \cellcolor[RGB]{219,231,244}-- & \cellcolor[RGB]{219,231,244}\phantom{0}\textbf{9.6} & \cellcolor[RGB]{219,231,244}77.8 \\
    \cellcolor[RGB]{219,231,244}\textcolor[HTML]{0072B2}{$\rightarrow$}\,+ $\DTCprod$ & \cellcolor[RGB]{219,231,244}-- & \cellcolor[RGB]{219,231,244}18.2 & \cellcolor[RGB]{219,231,244}75.1 & \cellcolor[RGB]{219,231,244}-- & \cellcolor[RGB]{219,231,244}16.6 & \cellcolor[RGB]{219,231,244}\textbf{79.6} & \cellcolor[RGB]{219,231,244}-- & \cellcolor[RGB]{219,231,244}19.5 & \cellcolor[RGB]{219,231,244}\textbf{78.7} & \cellcolor[RGB]{219,231,244}-- & \cellcolor[RGB]{219,231,244}17.1 & \cellcolor[RGB]{219,231,244}\textbf{80.5} \\
    Verb. Conf. & 75.7 & 44.5 & \textbf{83.9} & 60.6 & 44.3 & \textbf{80.8} & 41.8 & 42.9 & \textbf{85.3} & 43.9 & 43.9 & 79.2 \\
    \cellcolor[RGB]{219,231,244}\textcolor[HTML]{0072B2}{$\rightarrow$}\,+ $\DTClin$ & \cellcolor[RGB]{219,231,244}-- & \cellcolor[RGB]{219,231,244}\textbf{14.0} & \cellcolor[RGB]{219,231,244}73.7 & \cellcolor[RGB]{219,231,244}-- & \cellcolor[RGB]{219,231,244}\textbf{13.2} & \cellcolor[RGB]{219,231,244}73.7 & \cellcolor[RGB]{219,231,244}-- & \cellcolor[RGB]{219,231,244}\textbf{11.4} & \cellcolor[RGB]{219,231,244}74.3 & \cellcolor[RGB]{219,231,244}-- & \cellcolor[RGB]{219,231,244}\textbf{10.7} & \cellcolor[RGB]{219,231,244}78.5 \\
    \cellcolor[RGB]{219,231,244}\textcolor[HTML]{0072B2}{$\rightarrow$}\,+ $\DTCprod$ & \cellcolor[RGB]{219,231,244}-- & \cellcolor[RGB]{219,231,244}20.6 & \cellcolor[RGB]{219,231,244}78.4 & \cellcolor[RGB]{219,231,244}-- & \cellcolor[RGB]{219,231,244}18.4 & \cellcolor[RGB]{219,231,244}77.8 & \cellcolor[RGB]{219,231,244}-- & \cellcolor[RGB]{219,231,244}19.3 & \cellcolor[RGB]{219,231,244}80.5 & \cellcolor[RGB]{219,231,244}-- & \cellcolor[RGB]{219,231,244}16.6 & \cellcolor[RGB]{219,231,244}\textbf{81.4} \\
    Verb. TopK & 71.1 & 40.2 & \textbf{88.2} & 55.8 & 40.6 & \textbf{88.7} & 40.7 & 41.2 & \textbf{85.4} & 43.4 & 41.6 & \textbf{83.0} \\
    \cellcolor[RGB]{219,231,244}\textcolor[HTML]{0072B2}{$\rightarrow$}\,+ $\DTClin$ & \cellcolor[RGB]{219,231,244}-- & \cellcolor[RGB]{219,231,244}\textbf{36.4} & \cellcolor[RGB]{219,231,244}44.3 & \cellcolor[RGB]{219,231,244}-- & \cellcolor[RGB]{219,231,244}\textbf{35.2} & \cellcolor[RGB]{219,231,244}49.0 & \cellcolor[RGB]{219,231,244}-- & \cellcolor[RGB]{219,231,244}\textbf{22.4} & \cellcolor[RGB]{219,231,244}64.5 & \cellcolor[RGB]{219,231,244}-- & \cellcolor[RGB]{219,231,244}\textbf{21.9} & \cellcolor[RGB]{219,231,244}66.6 \\
    \cellcolor[RGB]{219,231,244}\textcolor[HTML]{0072B2}{$\rightarrow$}\,+ $\DTCprod$ & \cellcolor[RGB]{219,231,244}-- & \cellcolor[RGB]{219,231,244}46.7 & \cellcolor[RGB]{219,231,244}45.2 & \cellcolor[RGB]{219,231,244}-- & \cellcolor[RGB]{219,231,244}47.0 & \cellcolor[RGB]{219,231,244}50.2 & \cellcolor[RGB]{219,231,244}-- & \cellcolor[RGB]{219,231,244}32.9 & \cellcolor[RGB]{219,231,244}71.0 & \cellcolor[RGB]{219,231,244}-- & \cellcolor[RGB]{219,231,244}37.0 & \cellcolor[RGB]{219,231,244}69.0 \\
    Verb. PD & 75.3 & 33.5 & \textbf{75.5} & 62.6 & 34.1 & 69.6 & 41.0 & 31.1 & \textbf{77.3} & 43.7 & 34.9 & 72.3 \\
    \cellcolor[RGB]{219,231,244}\textcolor[HTML]{0072B2}{$\rightarrow$}\,+ $\DTClin$ & \cellcolor[RGB]{219,231,244}-- & \cellcolor[RGB]{219,231,244}\textbf{18.8} & \cellcolor[RGB]{219,231,244}63.0 & \cellcolor[RGB]{219,231,244}-- & \cellcolor[RGB]{219,231,244}\textbf{16.8} & \cellcolor[RGB]{219,231,244}72.0 & \cellcolor[RGB]{219,231,244}-- & \cellcolor[RGB]{219,231,244}\textbf{17.9} & \cellcolor[RGB]{219,231,244}68.4 & \cellcolor[RGB]{219,231,244}-- & \cellcolor[RGB]{219,231,244}\textbf{15.3} & \cellcolor[RGB]{219,231,244}77.3 \\
    \cellcolor[RGB]{219,231,244}\textcolor[HTML]{0072B2}{$\rightarrow$}\,+ $\DTCprod$ & \cellcolor[RGB]{219,231,244}-- & \cellcolor[RGB]{219,231,244}31.7 & \cellcolor[RGB]{219,231,244}67.3 & \cellcolor[RGB]{219,231,244}-- & \cellcolor[RGB]{219,231,244}29.4 & \cellcolor[RGB]{219,231,244}\textbf{75.5} & \cellcolor[RGB]{219,231,244}-- & \cellcolor[RGB]{219,231,244}29.2 & \cellcolor[RGB]{219,231,244}76.3 & \cellcolor[RGB]{219,231,244}-- & \cellcolor[RGB]{219,231,244}29.3 & \cellcolor[RGB]{219,231,244}\textbf{80.6} \\
    \bottomrule
  \end{tabular}
\end{table}


\begin{table}[H]
  \centering
  \caption{Verbalized vs.\ w/ DTC on Qwen3-30B-A3B.}
  \label{tab:verbalized-dtc-qwen30b}
  \footnotesize
  \setlength{\tabcolsep}{1.0pt}%
  \renewcommand{\arraystretch}{1.05}%
  \begin{tabular}{@{}wl{4.6em}@{\hspace{2.0pt}}wc{2.80em}wc{2.80em}@{\hspace{2.0pt}}wc{2.80em}wc{2.80em}@{\hspace{2.0pt}}wc{2.80em}wc{2.80em}@{\hspace{2.0pt}}wc{2.80em}wc{2.80em}@{}}
    \toprule
    \multirow{2}{*}{\centering\textbf{Method}} & \multicolumn{2}{c}{\textbf{AIME24}} & \multicolumn{2}{c}{\textbf{AIME25}} & \multicolumn{2}{c}{\textbf{HMMT25}} & \multicolumn{2}{c}{\textbf{HMMT26}} \\
    \cmidrule(lr){2-3}\cmidrule(lr){4-5}\cmidrule(lr){6-7}\cmidrule(lr){8-9}
    & \textbf{ECE}$\downarrow$ & \textbf{AUC}$\uparrow$ & \textbf{ECE}$\downarrow$ & \textbf{AUC}$\uparrow$ & \textbf{ECE}$\downarrow$ & \textbf{AUC}$\uparrow$ & \textbf{ECE}$\downarrow$ & \textbf{AUC}$\uparrow$ \\
    \midrule
    Verb. Conf. & 44.5 & 83.9 & 44.3 & 80.8 & 42.9 & 85.3 & 43.9 & 79.2 \\
    \cellcolor[RGB]{219,231,244}w/ DTC & \cellcolor[RGB]{219,231,244}{17.8} & \cellcolor[RGB]{219,231,244}{87.0} & \cellcolor[RGB]{219,231,244}{15.7} & \cellcolor[RGB]{219,231,244}{83.8} & \cellcolor[RGB]{219,231,244}{13.0} & \cellcolor[RGB]{219,231,244}{87.9} & \cellcolor[RGB]{219,231,244}{17.4} & \cellcolor[RGB]{219,231,244}{85.9} \\
    Verb. TopK & 40.2 & 88.2 & 40.6 & 88.7 & 41.2 & 85.4 & 41.6 & 83.0 \\
    \cellcolor[RGB]{219,231,244}w/ DTC & \cellcolor[RGB]{219,231,244}{11.1} & \cellcolor[RGB]{219,231,244}{83.0} & \cellcolor[RGB]{219,231,244}{10.4} & \cellcolor[RGB]{219,231,244}{84.9} & \cellcolor[RGB]{219,231,244}{\phantom{0}5.4} & \cellcolor[RGB]{219,231,244}{86.5} & \cellcolor[RGB]{219,231,244}{\phantom{0}6.1} & \cellcolor[RGB]{219,231,244}{86.4} \\
    Verb. PD & 33.5 & 75.5 & 34.1 & 69.6 & 31.1 & 77.3 & 34.9 & 72.3 \\
    \cellcolor[RGB]{219,231,244}w/ DTC & \cellcolor[RGB]{219,231,244}{23.8} & \cellcolor[RGB]{219,231,244}{79.7} & \cellcolor[RGB]{219,231,244}{23.4} & \cellcolor[RGB]{219,231,244}{76.9} & \cellcolor[RGB]{219,231,244}{22.1} & \cellcolor[RGB]{219,231,244}{80.0} & \cellcolor[RGB]{219,231,244}{20.1} & \cellcolor[RGB]{219,231,244}{79.5} \\
    \bottomrule
  \end{tabular}
\end{table}

\begin{table}[H]
  \centering
  \caption{Verbalized vs.\ w/ DTC on Qwen3-4B.}
  \label{tab:verbalized-dtc-qwen4b}
  \footnotesize
  \setlength{\tabcolsep}{1.0pt}%
  \renewcommand{\arraystretch}{1.05}%
  \begin{tabular}{@{}wl{4.6em}@{\hspace{2.0pt}}wc{2.80em}wc{2.80em}@{\hspace{2.0pt}}wc{2.80em}wc{2.80em}@{\hspace{2.0pt}}wc{2.80em}wc{2.80em}@{\hspace{2.0pt}}wc{2.80em}wc{2.80em}@{}}
    \toprule
    \multirow{2}{*}{\centering\textbf{Method}} & \multicolumn{2}{c}{\textbf{AIME24}} & \multicolumn{2}{c}{\textbf{AIME25}} & \multicolumn{2}{c}{\textbf{HMMT25}} & \multicolumn{2}{c}{\textbf{HMMT26}} \\
    \cmidrule(lr){2-3}\cmidrule(lr){4-5}\cmidrule(lr){6-7}\cmidrule(lr){8-9}
    & \textbf{ECE}$\downarrow$ & \textbf{AUC}$\uparrow$ & \textbf{ECE}$\downarrow$ & \textbf{AUC}$\uparrow$ & \textbf{ECE}$\downarrow$ & \textbf{AUC}$\uparrow$ & \textbf{ECE}$\downarrow$ & \textbf{AUC}$\uparrow$ \\
    \midrule
    Verb. Conf. & 45.6 & 80.8 & 45.9 & 85.6 & 45.6 & 81.9 & 45.1 & 79.8 \\
    \cellcolor[RGB]{219,231,244}w/ DTC & \cellcolor[RGB]{219,231,244}{30.8} & \cellcolor[RGB]{219,231,244}{82.0} & \cellcolor[RGB]{219,231,244}{29.4} & \cellcolor[RGB]{219,231,244}{87.0} & \cellcolor[RGB]{219,231,244}{31.0} & \cellcolor[RGB]{219,231,244}{81.6} & \cellcolor[RGB]{219,231,244}{31.1} & \cellcolor[RGB]{219,231,244}{81.2} \\
    Verb. TopK & 47.0 & 68.3 & 47.0 & 75.3 & 46.8 & 74.0 & 45.5 & 77.3 \\
    \cellcolor[RGB]{219,231,244}w/ DTC & \cellcolor[RGB]{219,231,244}{33.7} & \cellcolor[RGB]{219,231,244}{68.6} & \cellcolor[RGB]{219,231,244}{32.8} & \cellcolor[RGB]{219,231,244}{75.5} & \cellcolor[RGB]{219,231,244}{34.3} & \cellcolor[RGB]{219,231,244}{73.8} & \cellcolor[RGB]{219,231,244}{30.9} & \cellcolor[RGB]{219,231,244}{76.9} \\
    Verb. PD & 46.9 & 69.9 & 48.1 & 73.7 & 43.9 & 75.5 & 42.9 & 70.9 \\
    \cellcolor[RGB]{219,231,244}w/ DTC & \cellcolor[RGB]{219,231,244}{33.6} & \cellcolor[RGB]{219,231,244}{72.8} & \cellcolor[RGB]{219,231,244}{35.7} & \cellcolor[RGB]{219,231,244}{76.6} & \cellcolor[RGB]{219,231,244}{30.3} & \cellcolor[RGB]{219,231,244}{77.2} & \cellcolor[RGB]{219,231,244}{29.9} & \cellcolor[RGB]{219,231,244}{73.4} \\
    \bottomrule
  \end{tabular}
\end{table}

\section{Additional Analyses}
\label{app:additional_analyses}

\subsection{Black-box Auxiliary Size}
\label{app:dual-aux}

\begin{figure}[H]
  \centering
  \begin{subfigure}[t]{0.32\textwidth}
    \centering
    \includegraphics[width=\linewidth]{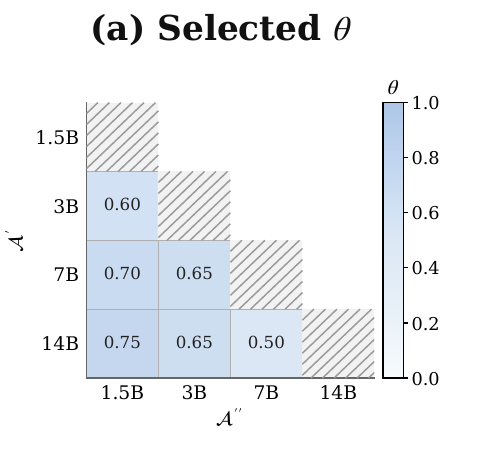}
    \label{fig:clin-theta-blackbox-a}
  \end{subfigure}\hfill
  \begin{subfigure}[t]{0.32\textwidth}
    \centering
    \includegraphics[width=\linewidth]{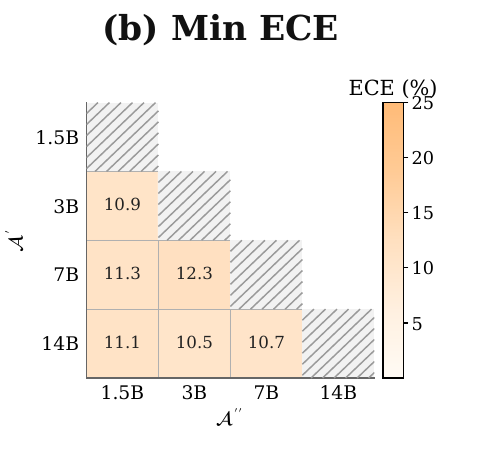}
    \label{fig:clin-theta-blackbox-b}
  \end{subfigure}\hfill
  \begin{subfigure}[t]{0.32\textwidth}
    \centering
    \includegraphics[width=\linewidth]{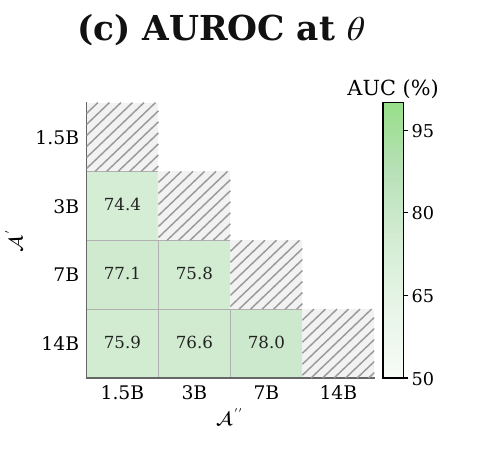}
    \label{fig:clin-theta-blackbox-c}
  \end{subfigure}
  \caption{Black-box sensitivity of $\DTClin$ to $\theta$ and auxiliary size. We search $\theta$ in steps of $0.05$ and report the $\theta$ (a) with the lowest ECE (b) and its AUROC (c). Cells without a finished probe are left blank.}
  \label{fig:clin-theta-blackbox}
\end{figure}

For dual-auxiliary \(\DTClin\), DeepSeek-V3.2 generates the trajectories and both auxiliaries are Qwen2.5 models~(\Cref{fig:clin-theta-blackbox}).
On AIME25 we search \(\theta\) in steps of \(0.05\) and report the \(\theta\) with the lowest ECE.
The lowest ECE is similar across auxiliary pairs.

\FloatBarrier

\subsection{Black-box Threshold Curves}
\label{app:blackbox-theta-curves}

\begin{figure}[H]
  \centering
  \includegraphics[width=0.58\textwidth]{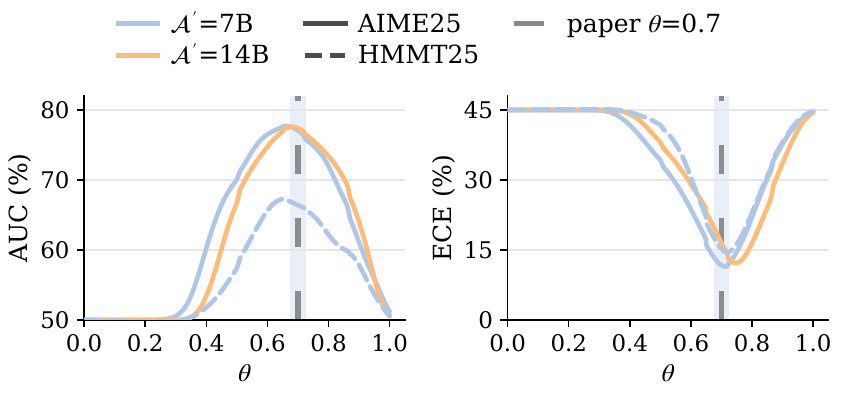}
  \caption{$\DTClin$ sensitivity to $\theta$ with $\gA''$ fixed at 1.5B (DeepSeek-V3.2 generating; solid: AIME25; dashed: HMMT25). The shaded band marks the black-box operating point $\theta{=}0.70$.}
  \label{fig:clin-theta-blackbox-curves}
\end{figure}

With \(\gA''\) fixed at 1.5B, we vary \(\theta\) for black-box \(\DTClin\) on DeepSeek-V3.2 trajectories~(\Cref{fig:clin-theta-blackbox-curves}).
The main experiments use \(\theta{=}0.70\), where ECE is still low.

\FloatBarrier

\subsection{Count and Ratio}
\label{app:count-vs-ratio}

\begin{figure}[H]
  \centering
  \includegraphics[width=0.66\textwidth]{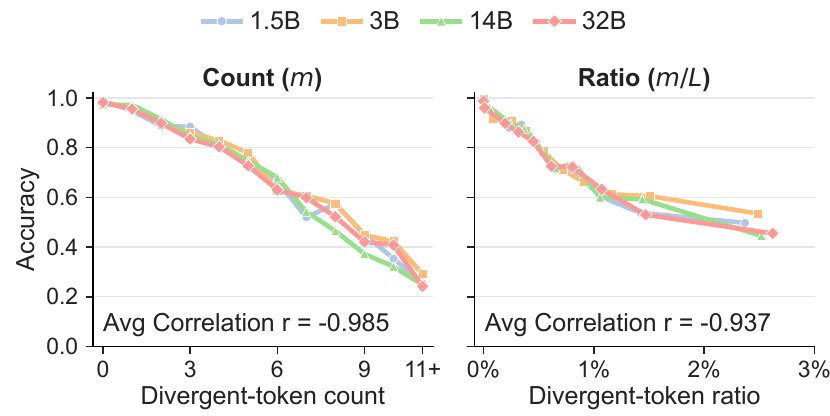}
  \caption{Divergent-token count and ratio against path accuracy, single-auxiliary Qwen2.5-7B on MATH-500.
  Each curve is one auxiliary at the threshold used in Figure~\ref{fig:divergent-token-count-vs-acc}(c).}
  \label{fig:count-vs-ratio}
\end{figure}

Longer reasoning traces are often less accurate, and recent work links that drop to overthinking~\citep{Ghosal-2025-Thinking-More,Gema-2025-Inverse-Scaling}.
We check whether the divergent-token count is only a proxy for this length effect.
Let \(L\) be the number of reasoning tokens on a path.
The count grows with \(L\), so the two panels of~\Cref{fig:count-vs-ratio} plot:
\begin{equation}
  m=\lvert T_{\rm div}(\theta)\rvert,
  \qquad
  \frac{m}{L}
  =\frac{\lvert T_{\rm div}(\theta)\rvert}{L}.
  \label{eq:count-ratio}
\end{equation}
These runs use the single-auxiliary setting of~\Cref{fig:divergent-token-count-vs-acc}(c): MATH-500, Qwen2.5-7B, with auxiliaries of 1.5B, 3B, 14B, and 32B, each at the threshold used there.

On the left of~\Cref{fig:count-vs-ratio}, accuracy falls from about \(0.97\)--\(0.99\) at a count of zero to about \(0.24\)--\(0.29\) once the count reaches \(11\) or more. On the right, we group the same trajectories into ten quantile bins of the ratio. Accuracy still falls from about \(0.97\)--\(0.99\) in the lowest bin to about \(0.45\)--\(0.54\) in the highest. The correlations of the bin means are \(r=-0.985\) for the count and \(r=-0.937\) for the ratio. Dividing by path length does not remove the decline, so the count is not only a proxy for path length.

\FloatBarrier

\subsection{Choice of Divergence Measure}
\label{app:divergence-ablation}

\begin{figure}[H]
  \centering
  \includegraphics[width=\textwidth]{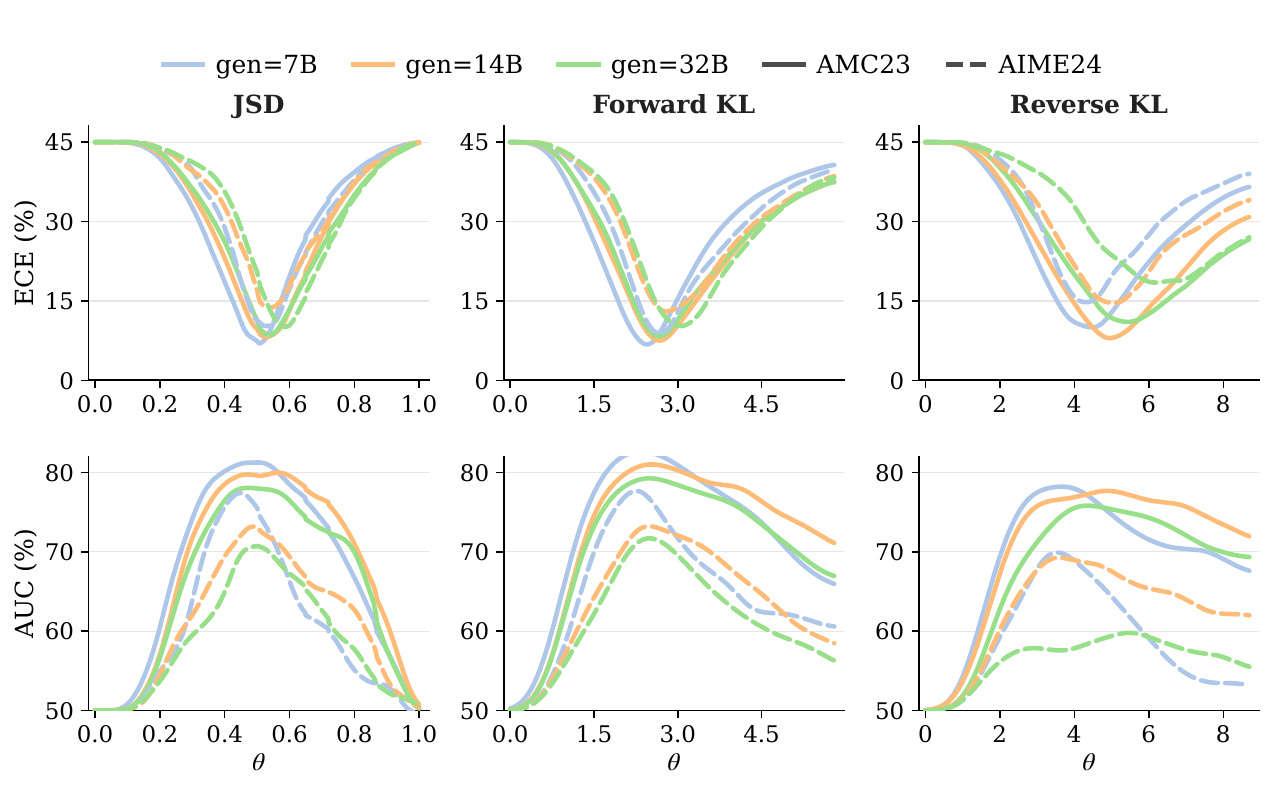}
  \caption{Divergence choice for divergent-token selection.
  Each column is one disagreement score (JSD, forward KL, reverse KL)
  for $\DTClin$ with auxiliary fixed at 1.5B
  and generating models $\in\{7,14,32\}$B (solid: AMC23; dashed: AIME24).
  The top row shows ECE (\%) and the bottom row AUROC (\%) versus $\theta$.}
  \label{fig:divergence-ablation}
\end{figure}

\begin{figure}[H]
  \centering
  \includegraphics[width=\textwidth]{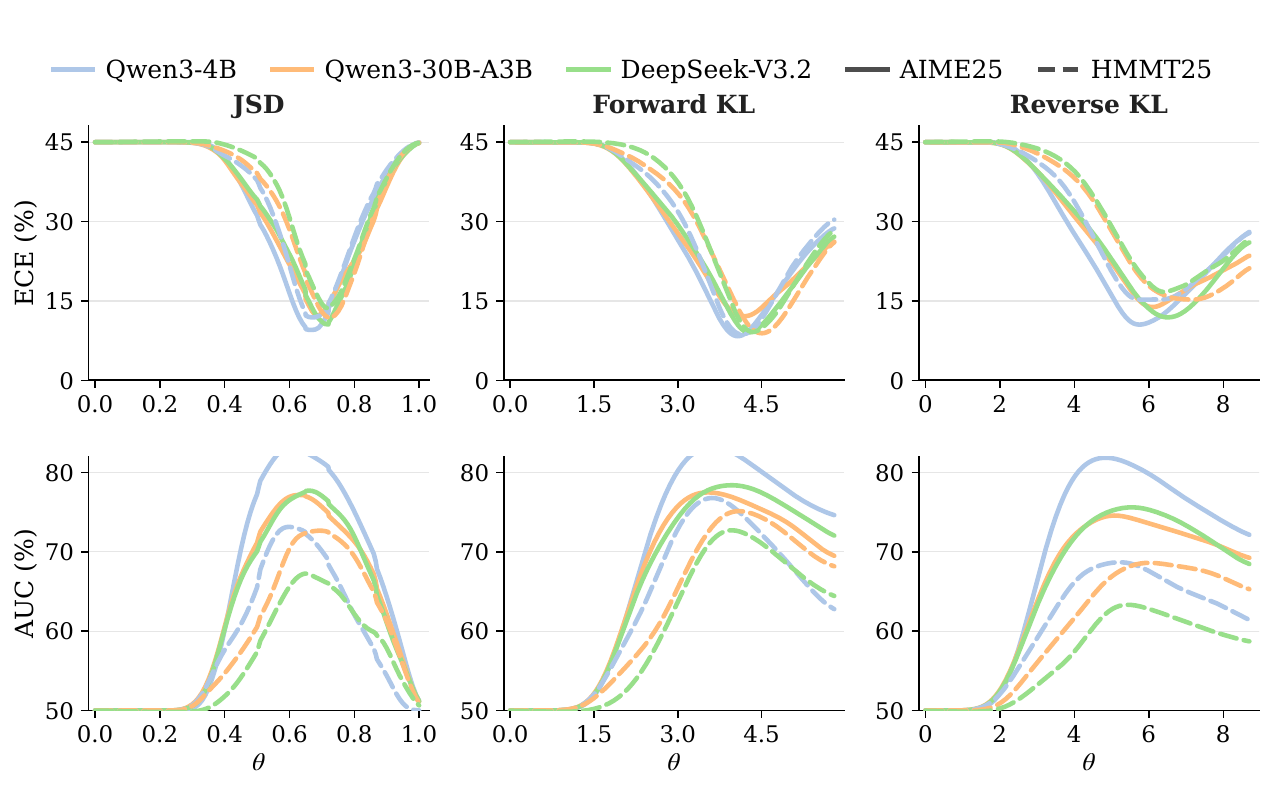}
  \caption{Divergence choice for black-box divergent-token selection.
  Each column is one disagreement score (JSD, forward KL, reverse KL)
  for $\DTClin$ with auxiliaries fixed at Qwen2.5-7B and Qwen2.5-1.5B
  and generators Qwen3-4B, Qwen3-30B-A3B, and DeepSeek-V3.2
  (solid: AIME25; dashed: HMMT25).
  The top row shows ECE (\%) and the bottom row AUROC (\%) versus $\theta$.}
  \label{fig:divergence-ablation-blackbox}
\end{figure}

We keep the count estimator and the auxiliaries fixed, and compare JSD with forward KL and reverse KL~(\Cref{fig:divergence-ablation,fig:divergence-ablation-blackbox}).
White-box runs use a 1.5B auxiliary and generators of 7B, 14B, and 32B. Black-box runs use Qwen2.5-7B and Qwen2.5-1.5B.
ECE against \(\theta\) is U-shaped for all three scores, and AUROC rises where ECE falls.
KL needs a larger \(\theta\) than JSD, and that \(\theta\) is less stable from model to model.
JSD is steadier near the \(\theta\) used in the main experiments.

\subsection{Threshold and Saturation Sensitivity}
\label{app:threshold-sensitivity}
\label{app:clin-n-ablation}

\begin{figure}[H]
  \centering
  \includegraphics[width=\textwidth]{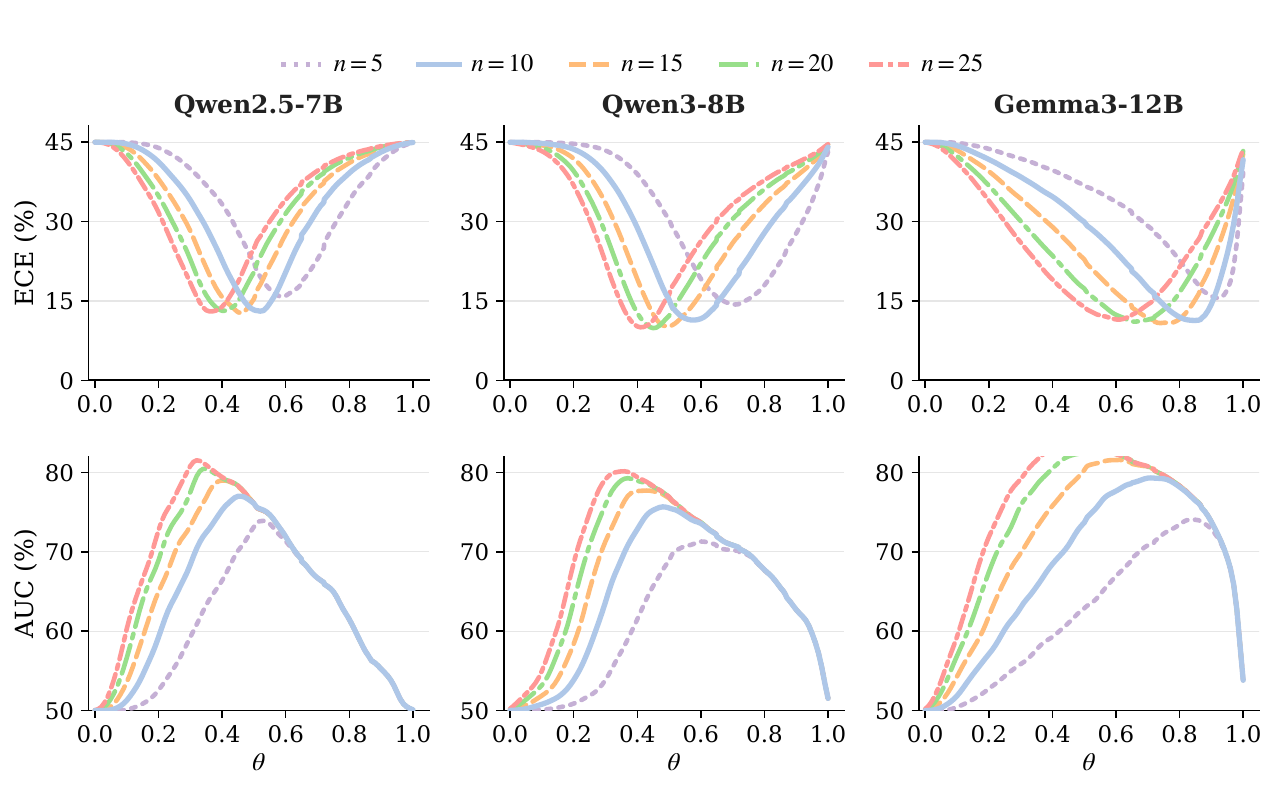}
  \caption{Saturation length $n$ for $\DTClin$.
  The top row shows ECE (\%) and the bottom row AUROC (\%) versus $\theta$ at step $0.01$,
  averaged over MATH-500, AMC23, AIME24, and AIME25.}
  \label{fig:clin-n-ablation}
\end{figure}

In the white-box setting, \Cref{fig:clin-n-ablation} varies \(\theta\) and the saturation length \(n\) of \(\DTClin\).
Each panel uses the smallest generator in its family: Qwen2.5-7B, Qwen3-8B, or Gemma3-12B. The curves average MATH-500, AMC23, AIME24, and AIME25.
The ECE minimum moves with the family, but the \(\theta\) values from the main text are still where ECE is low, and the comparison across \(n\) supports \(n=10\).

\subsection{Black-box Saturation Length}
\label{app:blackbox-saturation}

\begin{figure}[H]
  \centering
  \includegraphics[width=\textwidth]{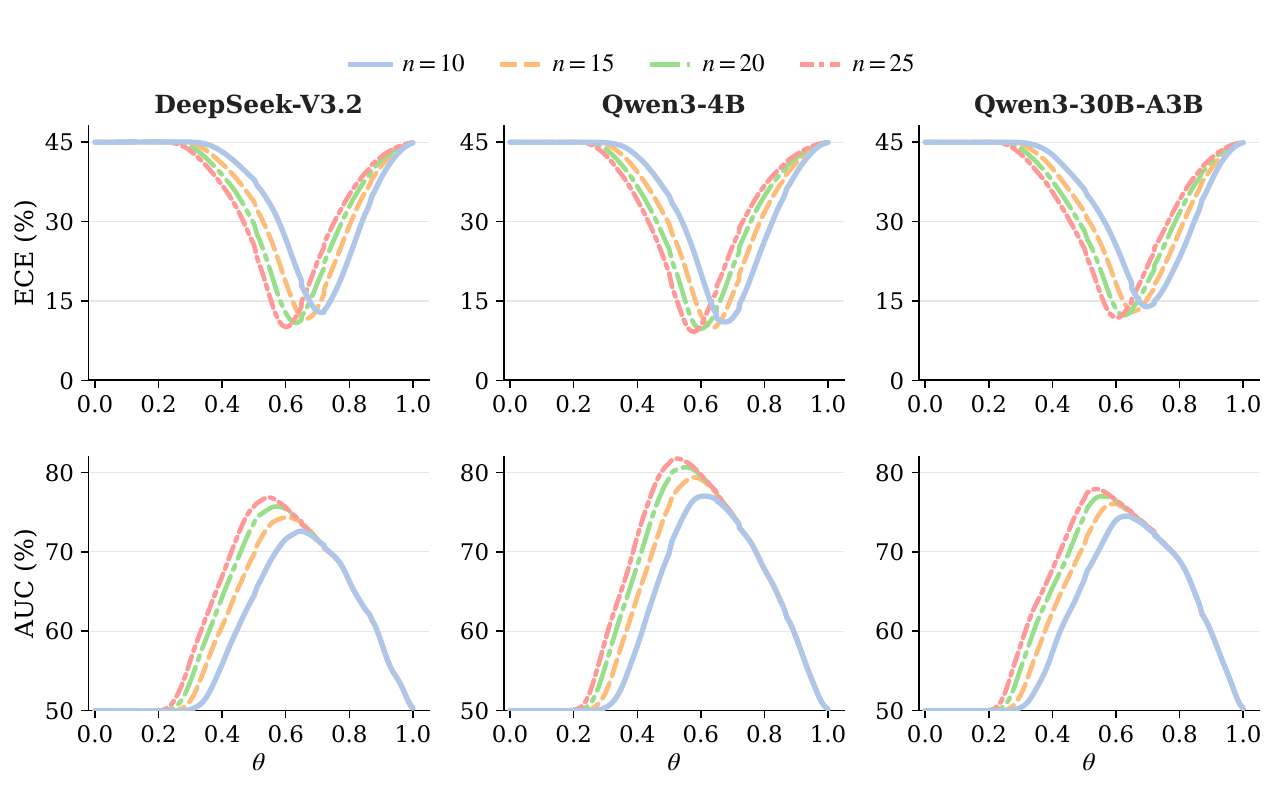}
  \caption{Saturation length $n$ for black-box $\DTClin$.
  Each column is one generator with Qwen2.5 auxiliaries $7$B and $1.5$B.
  The top row shows ECE (\%) and the bottom row AUROC (\%) versus $\theta$ at step $0.01$,
  averaged over AIME24, AIME25, and HMMT25.}
  \label{fig:clin-n-ablation-blackbox}
\end{figure}

Black-box \(\DTClin\) uses DeepSeek-V3.2, Qwen3-4B, and Qwen3-30B-A3B, with Qwen2.5 auxiliaries of 7B and 1.5B~(\Cref{fig:clin-n-ablation-blackbox}).
Averaged over AIME24, AIME25, and HMMT25, these curves also support \(n=10\).

\subsection{Offset $k$ for $\DTCprod$}
\label{app:dtcprod-k-ablation}

\begin{figure}[H]
  \centering
  \includegraphics[width=\textwidth]{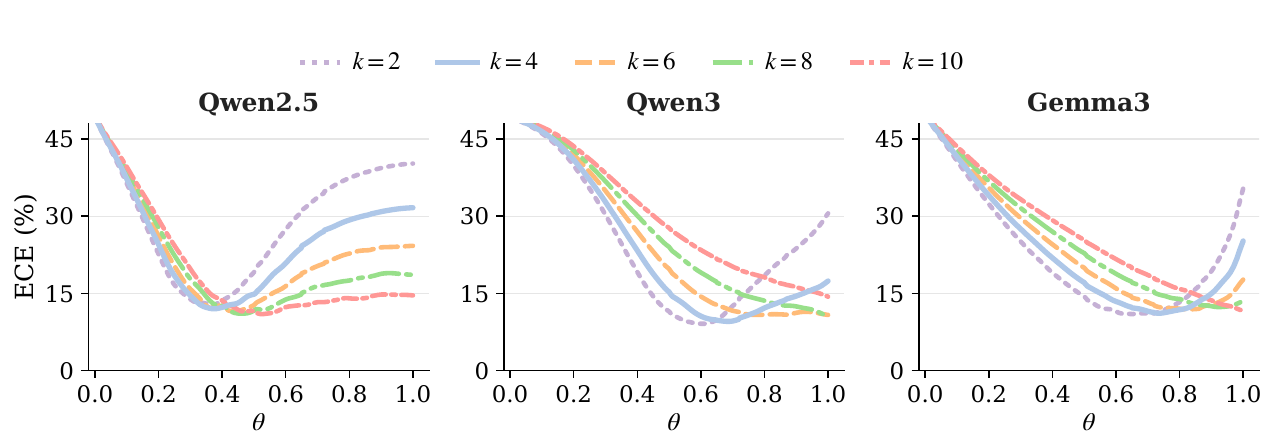}
  \caption{Offset $k$ for $\DTCprod$.
  Each column is the smallest white-box generator in its family
  (Qwen2.5-7B, Qwen3-8B, Gemma3-12B) with the family auxiliary.
  ECE (\%) is shown versus $\theta$ at step $0.01$,
  averaged over MATH-500, AMC23, AIME24, and AIME25.}
  \label{fig:dtcprod-k-ablation}
\end{figure}

\Cref{fig:dtcprod-k-ablation} varies \(k\) in \(\DTCprod\) on the smallest white-box generator in each family.
ECE is averaged over MATH-500, AMC23, AIME24, and AIME25.
ECE against \(\theta\) stays U-shaped, and the minimum moves with \(k\) and with the family.
At the \(\theta\) used in the main text, \(k=4\) still gives low ECE, so we keep \(k=4\).

\FloatBarrier

\FloatBarrier

\FloatBarrier

\FloatBarrier

\end{document}